\documentclass{article}

 \usepackage[preprint]{neurips_2026}

\usepackage[utf8]{inputenc} 
\usepackage[T1]{fontenc}    
\usepackage{hyperref}       
\usepackage{url}            
\usepackage{booktabs}       
\usepackage{amsfonts}       
\usepackage{nicefrac}       
\usepackage{microtype}      
\usepackage{xcolor}         

\definecolor{linkblue}{HTML}{4682B4}
\definecolor{linkred}{HTML}{C00000}
\hypersetup{
    colorlinks=true,
    linkcolor=linkblue,
    citecolor=linkblue,
    urlcolor=linkblue,
}

\usepackage{graphicx}
\usepackage{subcaption}
\usepackage{latexsym}
\usepackage{multicol}
\usepackage{multirow}
\usepackage{inconsolata}
\usepackage{titlesec}
\usepackage{xspace}
\usepackage{tcolorbox}
\usepackage{fvextra}
\usepackage{amsmath}
\usepackage{amssymb}
\usepackage{mathtools}
\usepackage{amsthm}
\usepackage{algorithm}
\usepackage{algorithmic}
\usepackage[capitalise]{cleveref}
\usepackage[textsize=tiny]{todonotes}
\usepackage{wrapfig}

\newcommand{\methodName}{{\textsc{SPoT}}\xspace}

\newcommand{\pos}[1]{\scriptsize\textcolor{green!50!black}{$\uparrow$#1}}
\newcommand{\negat}[1]{\scriptsize\textcolor{red!70!black}{$\downarrow$#1}}

\newcommand{\modify}[1]{#1} 

\tcbuselibrary{breakable, skins}

\crefname{section}{Sec.}{Secs.}
\crefname{figure}{Fig.}{Figs.}
\crefname{equation}{Eq.}{Eqs.}
\crefname{table}{Tab.}{Tabs.}

\Crefname{section}{Section}{Sections}
\Crefname{figure}{Figure}{Figures}
\Crefname{equation}{Equation}{Equations}
\Crefname{table}{Table}{Tables}

\newcommand{\rednum}[3]{{\hypersetup{linkcolor=linkred}#1#2#3}}
\crefformat{figure}{Fig.~\rednum{#2}{#1}{#3}}
\Crefformat{figure}{Figure~\rednum{#2}{#1}{#3}}
\crefmultiformat{figure}{Figs.~\rednum{#2}{#1}{#3}}{ and~\rednum{#2}{#1}{#3}}{, \rednum{#2}{#1}{#3}}{ and~\rednum{#2}{#1}{#3}}
\Crefmultiformat{figure}{Figures~\rednum{#2}{#1}{#3}}{ and~\rednum{#2}{#1}{#3}}{, \rednum{#2}{#1}{#3}}{ and~\rednum{#2}{#1}{#3}}
\crefformat{table}{Tab.~\rednum{#2}{#1}{#3}}
\Crefformat{table}{Table~\rednum{#2}{#1}{#3}}
\crefmultiformat{table}{Tabs.~\rednum{#2}{#1}{#3}}{ and~\rednum{#2}{#1}{#3}}{, \rednum{#2}{#1}{#3}}{ and~\rednum{#2}{#1}{#3}}
\Crefmultiformat{table}{Tables~\rednum{#2}{#1}{#3}}{ and~\rednum{#2}{#1}{#3}}{, \rednum{#2}{#1}{#3}}{ and~\rednum{#2}{#1}{#3}}
\crefformat{section}{Sec.~\rednum{#2}{#1}{#3}}
\Crefformat{section}{Section~\rednum{#2}{#1}{#3}}
\crefmultiformat{section}{Secs.~\rednum{#2}{#1}{#3}}{ and~\rednum{#2}{#1}{#3}}{, \rednum{#2}{#1}{#3}}{ and~\rednum{#2}{#1}{#3}}
\Crefmultiformat{section}{Sections~\rednum{#2}{#1}{#3}}{ and~\rednum{#2}{#1}{#3}}{, \rednum{#2}{#1}{#3}}{ and~\rednum{#2}{#1}{#3}}

\theoremstyle{plain}

\theoremstyle{definition}

\theoremstyle{remark}

\title{Surgical Post-Training: Proximal On-Policy Distillation for Reasoning with Knowledge Retention}

\author{%
  Wenye Lin\\
  The University of Hong Kong\\
  \texttt{linius@connect.hku.hk} \\
  \And
  Kai Han\thanks{Corresponding author.} \\
  The University of Hong Kong \\
  \texttt{kaihanx@hku.hk} \\
}

\date{}

\begin{document}

\maketitle

\begin{abstract}
\modify{Injecting new reasoning knowledge into Large Language Models (LLMs) via post-training often induces catastrophic forgetting. Recent studies emphasize the importance of on-policy data but suggest that KL-divergence fails to mitigate forgetting. In contrast, we show, both analytically and empirically, that the KL-constrained reward formulation actually plays a critical role in retaining knowledge during post-training. This motivates our Surgical Post-Training (\methodName), a proximal on-policy distillation framework designed to optimize reasoning efficiently while preserving prior knowledge.} \methodName consists of (1) a data rectification pipeline employing an Oracle to surgically correct erroneous steps via minimal edits, generating proximal on-policy data; and (2) a reward-based binary cross-entropy objective \modify{essential for enhancing reasoning and mitigating forgetting.}
Empirically, with only 4k rectified math pairs, \methodName improves Qwen3-8B's accuracy by 6.2\% on average across in-domain and out-of-domain tasks, requiring merely 16-minute model training on 8$\times$ H800 GPUs. Moreover, \methodName provides a superior initialization for subsequent reinforcement learning, significantly elevating the performance ceiling. 
Code: \url{https://github.com/Visual-AI/SPoT}

\end{abstract}




\section{Introduction}

While pretraining builds the foundation of Large Language Models (LLMs)~\citep{achiam2023gpt,comanici2025gemini}, post-training remains critical for eliciting specific competencies, such as mathematical reasoning~\citep{shao2024deepseekmath}, code generation~\citep{anthropic2024claude}, and human alignment~\citep{ouyang2022training,zhou2023instruction}. Post-training for reasoning primarily follows two paradigms: Supervised Fine-Tuning (SFT) and Reinforcement Learning (RL). \modify{While SFT benefits from strong supervision, it induces catastrophic forgetting when injecting new reasoning capabilities~\citep{chu2025sft,huan2025does}. On-policy RL methods like Group Relative Policy Optimization (GRPO) \citep{shao2024deepseekmath} avoid forgetting and excel on tasks with verifiable outcomes, yet they rely on the base model's ability to sample correct paths, making them less suited for injecting new knowledge that expands fundamental reasoning boundaries~\citep{yue2025does}.}

\modify{This raises a critical question: \textbf{How can we effectively inject external reasoning knowledge into LLMs while mitigating catastrophic forgetting?} Recent studies reveal that \textit{on-policy data} is crucial for mitigating catastrophic forgetting, as it prevents the drastic distributional shifts inherent in off-policy SFT~\citep{chen2025retaining,shenfeld2025rl}. Motivated by this, a growing body of work explores on-policy distillation \citep{agarwal2024gkd,lu2025onpolicydistillation,yang2025qwen3,xiao2026mimo}, where the teacher model provides token-level distributions as dense supervision for the student's trajectories. While effective, this approach requires identical tokenizers and white-box teachers, precluding cross-family distillation or the use of proprietary models. To circumvent this constraint, we relax the white-box requirement and instead introduce a data rectification pipeline that generates proximal on-policy data without requiring the teacher's token distributions. Specifically, we employ an Oracle (e.g., a teacher model) to surgically correct erroneous student outputs via minimal edits, yielding positive samples that remain proximal to the model's original distribution.}

\textbf{However, we find that data proximity is necessary but insufficient.}\quad Empirically, SFT on this proximal data \textit{still} incurs forgetting, whereas Direct Preference Optimization (DPO)~\citep{rafailov2023direct} does not. To understand the mechanism, we employ a Reward-SFT baseline, a method integrating the DPO reward formulation into SFT. We observe that Reward-SFT effectively retains prior knowledge, confirming that the KL-constraint in the reward definition is the decisive factor: it ``tethers'' the updated policy to the reference model. 

Beyond preserving prior knowledge, the model must also achieve superior gains for in-domain reasoning tasks. We identify two failure modes: (1) \textit{the ``pull-up'' effect}~\citep{renlearning} inherent in positive-only training (e.g., SFT), which inadvertently raises the likelihood of erroneous responses alongside correct ones; and (2) \textit{the inadequacy of relative ranking} in DPO for reasoning with verifiable correctness.


To address these limitations, we propose Surgical Post-Training (\methodName), a paradigm designed to optimize reasoning efficiently while preserving prior knowledge. \methodName integrates our data rectification pipeline with a reward-based binary cross-entropy objective to maintain the ``tethering'' effect of the reference model. Unlike the relative ranking in DPO, our objective explicitly maximizes the likelihood of the rectifications while suppressing errors, showing superior gains in reasoning.

In this paper, we make the following contributions. \textit{Firstly}, we demonstrate that beyond on-policy data, implicit regularization is another critical factor in mitigating forgetting, showing that with identical data, SFT fails to generalize whereas DPO retains prior learned knowledge. \textit{Secondly}, we propose \methodName, a proximal on-policy distillation framework that mitigates catastrophic forgetting by synergizing minimal-edit data rectification with a reward-based objective; we identify the ``pull-up'' effect in positive-only training and the insufficiency of DPO for reasoning, and validate a binary classification objective that provides a denser signal. \textit{Thirdly}, \modify{we show that \methodName establishes a superior initialization for subsequent reinforcement learning: compared to applying GRPO directly, initializing GRPO with \methodName checkpoints substantially unlocks higher performance ceilings on in-domain math (+7.2\%) and Connect4 (+21.7\%), with better general instruction-following capabilities.} \textit{Lastly}, to strictly evaluate OOD reasoning without data contamination, we leverage GAMEBoT \citep{lin2025gamebot} to dynamically construct a \texttt{Connect4} dataset for reliable evaluation.

\begin{figure}[t]
  \centering
  \includegraphics[width=0.96\textwidth]{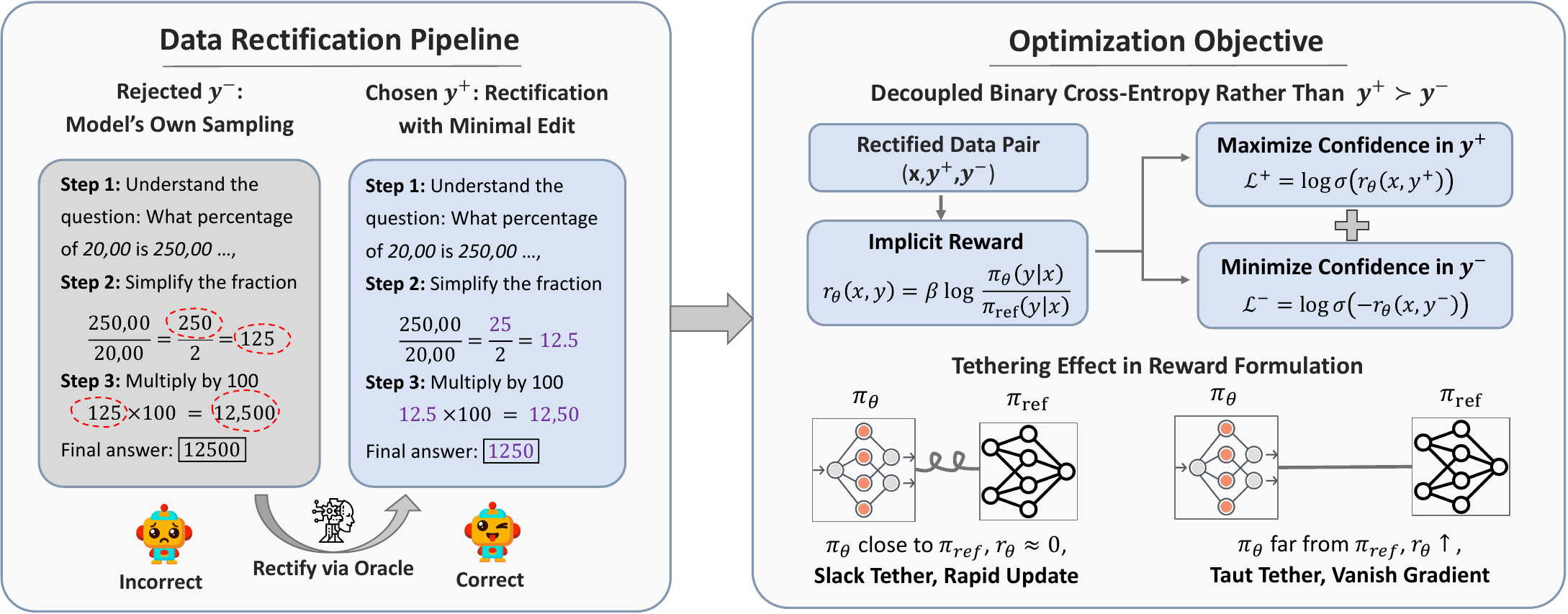}
  \caption{ \textbf{Illustration of \methodName.} Left: An Oracle surgically corrects reasoning errors, yielding positive samples proximal to the model's original distribution. Top-Right: Unlike DPO's relative ranking, we leverage an explicit classification loss that proves more effective for reasoning tasks. Bottom-Right: The tethering effect inherent in our reward definition substantially mitigates catastrophic forgetting.}
   \label{fig:illustration}
   \vspace{-6pt}
\end{figure}

\section{Data Rectification Pipeline}
\label{sec:pipeline}
Our \methodName framework consists of two core components: a \textit{data rectification pipeline} and a \textit{specialized optimization objective} (see~\cref{fig:illustration}). In this section, we detail the first component, which is designed to produce valid reasoning paths proximal to the model's distribution.

Standard SFT often utilizes offline datasets that diverge significantly from the model's current policy $\pi_\theta$, leading to catastrophic forgetting~\citep{chen2025retaining}. Conversely, on-policy methods rely entirely on the model's intrinsic probability of sampling correct reasoning paths, making them inefficient for hard reasoning tasks. For example, if a model's pass rate on a specific problem is 1\%, obtaining a single correct response requires roughly 100 rollouts in expectation, which is computationally expensive. Furthermore, for problems beyond the model's current capabilities, self-sampling fails to yield any training signal, making it impossible to learn from those instances~\citep{yu2025dapo}.

To bridge this gap, we propose a \textit{data rectification pipeline} to construct a contrastive dataset $\mathcal{D} = \{(x, y^-, y^+)\}$ where the positive response $y^+$ only performs necessary corrections to the erroneous logic of the negative response $y^-$ while preserving the model's original generation style and lexical structure. An example is shown in \cref{fig:illustration}. The pipeline consists of the following three stages:

\textbf{Error Elicitation.}\quad We first construct a dataset of model failures. Given a question $x$ from the source dataset, we sample a response $y^-$ from the current policy: $y^- \sim \pi_\theta(\cdot|x)$. We evaluate $y^-$ against the ground-truth answer. If the final answer is incorrect, we retain the pair $(x, y^-)$ for rectification. 

\textbf{Oracle-Guided Surgical Rectification.}\quad We employ an Oracle (e.g., a human expert or a teacher model) to perform surgical edits. The Oracle is provided with $y^-$ and optionally the ground truth, then instructed to only modify incorrect reasoning steps, while maintaining the original style. After filtering out any outputs with incorrect final answers, we obtain a rectified response $y^+$ that represents the ``nearest valid neighbor'' to the erroneous response $y^-$ in the semantic space. To enhance efficiency, we adopt an \textit{offline} setting (i.e., the Oracle is queried once per failure rather than at each training step). We provide the full prompts in Appendix \ref{appd:prompt}, and verify the framework's robustness to a weaker open-source Oracle and to prompt formulation in Appendix~\ref{appd:weak_oracle} and Appendix~\ref{appd:prompt_sensitivity}, respectively.

\label{sec:data}
\textbf{LCS Filtering.}\quad To strictly enforce the proximity of the rectified data to the student's distribution, we apply a structural constraint based on the Longest Common Subsequence (LCS) \citep{wagner1974string}. For every pair $(y^-, y^+)$, we calculate the change ratio $R_{LCS}$:

\begin{equation}
    R_{LCS}(y^-, y^+) = 1-\frac{|\text{LCS}(y^-, y^+)|}{|y^+|},
    \label{eq:change_ratio}
\end{equation}
where $|\cdot|$ denotes sequence length. We filter out samples where $R_{LCS} > \gamma$. We set $\gamma=0.6$ to maximize the retention of training samples while preserving downstream performance (see Appendix~\ref{appd:lcs_threshold} for the rationale and ablation). Appendix \ref{appd:change_ratio} shows the change ratio distributions.

In the resulting dataset, $y^-$ and $y^+$ in each sample share the majority of their token trajectories, diverging only at decision-critical parts. This is essential for our subsequent optimization, as it allows the gradients to focus on the divergent reasoning tokens.

\section{The Reward Is Secretly a Regularizer}
\label{sec:reward}
Building upon the proximal dataset constructed via the pipeline in \cref{sec:pipeline}, we demonstrate that data proximity alone is insufficient to prevent catastrophic forgetting. We analyze the learning dynamics of SFT versus reward-based methods, showing that the learning objective itself serves as an intrinsic regularizer against forgetting.
\subsection{Empirical Observation: The Regularization Gap}

We conduct controlled experiments using the English subset of DAPO-Math-17k \citep{yu2025dapo}. Leveraging our pipeline, we generate 4k contrastive pairs by employing Qwen3-8B as the base policy $\pi_\theta$ and Gemini 2.5 Pro as the Oracle for rectification.
We compare the following methods, all trained on $\mathcal{D} = \{(x, y^-, y^+)\}$, and evaluate OOD generalization using IFEval~\citep{zhou2023instruction}:
\textbf{SFT+.}\quad To distinguish it from standard fine-tuning on raw off-policy data, we refer to Supervised Fine-Tuning performed on the surgically rectified positive samples $y^+$ as SFT+, which minimizes the same negative log-likelihood as SFT:
\begin{equation}
    \mathcal{L}_{\text{SFT}} = - \mathbb{E}_{x, y^+ \sim \mathcal{D}} [\log \pi_\theta(y^+|x)]
\end{equation}

\textbf{DPO.}\quad DPO optimizes a policy by leveraging a relative ranking objective derived from an implicit reward $r_\theta(x, y)$. It is defined as the log-ratio between the policy $\pi_\theta$ and the frozen reference model $\pi_{\text{ref}}$ (initialized with the same parameters as $\pi_\theta$), scaled by a coefficient $\beta$:
\begin{equation}
    r_\theta(x, y) = \beta \log \frac{\pi_\theta(y|x)}{\pi_{\text{ref}}(y|x)} \label{eq:reward_def}
\end{equation}
DPO maximizes the margin between the chosen response $y^+$ and the rejected response $y^-$:
\begin{equation}
\hspace{-3pt}
    \mathcal{L}_{\text{DPO}} = - \mathbb{E}_{x, y^+, y^- \sim \mathcal{D}} [\log \sigma \! \left( r_\theta(x, y^+) \!-\! r_\theta(x, y^-) \right)] \label{eq:dpo_loss}
\end{equation}
where $\sigma(z) = 1/(1+e^{-z})$ is the sigmoid function.

\textbf{Reward-SFT.}\quad We introduce Reward-SFT as a control baseline to isolate the effect of the implicit reward from the influence of negative samples. Unlike SFT, which directly maximizes the token-level likelihood, this objective simply maximizes the probability of the chosen response $y^+$ being classified as ``correct'' under the implicit reward formulation:
 \begin{equation}
    \mathcal{L}_{\text{RW-SFT}} = - \mathbb{E}_{x, y^+ \sim \mathcal{D}} [\log \sigma \! \left( r_\theta(x, y^+) \right)] \label{eq:rewardsft_loss}
\end{equation}

\begin{wrapfigure}{r}{0.46\textwidth}
  \centering
  \vspace{-10pt} 
  \includegraphics[width=0.38\textwidth]{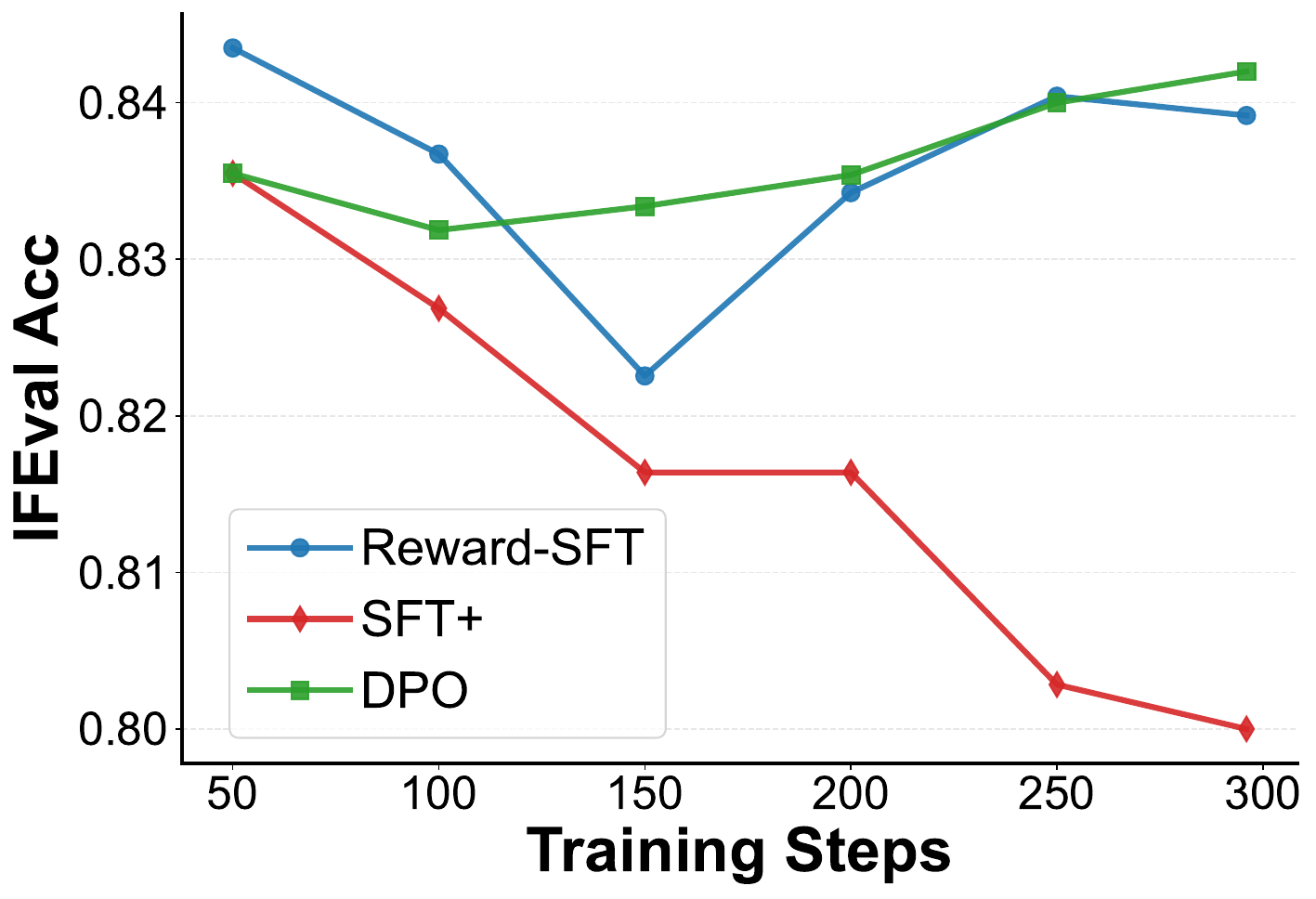}
  \caption{\textbf{IFEval acc results (avg@5).} SFT+ forgets instruction following ability, while reward-based methods do not. }
  \vspace{-17pt} 
  \label{fig:ifeval}
\end{wrapfigure}
Functionally, this treats the optimization as a binary classification task where the model learns to classify $y^+$ as ``high reward'' relative to the reference distribution. As illustrated in \cref{fig:ifeval}, we observe a significant divergence in generalization capability. SFT+ suffers from immediate catastrophic forgetting, with IFEval performance decaying monotonically as training progresses. DPO remains stable and even improves slightly. Crucially, similar to SFT+, Reward-SFT does not see negative data, yet it matches the OOD stability of DPO. This finding suggests that the resistance to forgetting stems from \emph{the implicit regularization inherent in the reward formulation}.

\subsection{Gradient Analysis: The Mechanics of Tethering}

To understand why Reward-SFT prevents forgetting while the standard SFT objective fails, we analyze the gradient dynamics of their respective loss functions. We show that the distinction between these two objectives lies entirely in \emph{the dynamic scaling coefficient of the gradient}.

The gradient for SFT with respect to the model parameters $\theta$ is:
\begin{equation}
\nabla_\theta \mathcal{L}_{\text{SFT}} = - \nabla_\theta \log \pi_\theta(y^+|x)
\label{eq:sft_gradient}
\end{equation}

Meanwhile, the gradient for Reward-SFT in \cref{eq:rewardsft_loss} includes a dynamic re-weighting coefficient determined by $r_\theta$ from \cref{eq:reward_def}:
\begin{equation}
\nabla_\theta \mathcal{L}_{\text{RW-SFT}} = - \frac{\partial \log \sigma(r)}{\partial r} \nabla_\theta r_\theta(x, y^+) = - \left(1 - \sigma(r_\theta(x, y^+))\right) \cdot \beta \nabla_\theta \log \pi_\theta(y^+|x)
\label{eq:reward_sft_gradient}
\end{equation}
Comparing \Cref{eq:sft_gradient,eq:reward_sft_gradient} reveals a fundamental difference in optimization pressure. In standard SFT, the gradient is scaled by a constant factor of $1$. Consequently, SFT applies uniform optimization pressure to all samples, regardless of the model's current competence. Even if the model has assigned high probability to the training sample (e.g., $p(y^+|x) \approx 0.99$), the SFT objective continues to force parameter updates to push the probability toward $1.0$. This unbounded optimization compels the model to overwrite its pre-trained features, resulting in significant distribution shift and loss of prior capabilities.

In contrast, Reward-SFT introduces an instance-dependent dynamic scaling coefficient $\lambda(x, y^+) = 1 - \sigma(r_\theta(x, y^+))$ (for simplicity, we assume $\beta = 0.1$ in our analysis). We term $\lambda$ the \textit{Elastic Tether}, as it modulates the ``tension'' of gradient updates. Functioning as an adaptive regularizer based on the implicit reward, it creates two distinct training modes. \textbf{Acquisition (slack tether):} when $\pi_\theta$ is close to $\pi_{\text{ref}}$, the reward $r_\theta \approx 0$ and $\lambda \approx 0.5$, allowing rapid adaptation to the preferred distribution. \textbf{Saturation (taut tether):} as the model aligns with the target and $r_\theta$ grows, $\lambda = 1 - \sigma(r_\theta) \to 0$, restricting the policy from deviating excessively from the reference.

Consider a sample where the model has achieved high confidence relative to the reference, yielding a reward of $r_\theta = 10$. The sigmoid function saturates, and the gradient scaling coefficient $\lambda$ becomes:
\begin{equation}
    \lambda = 1 - \sigma(10) = 1 - \frac{1}{1 + e^{-10}} \approx 4.5 \times 10^{-5} \nonumber
\end{equation}
In comparison to standard SFT, where the coefficient is fixed at $1.0$, the Elastic Tether tightens the constraint by a factor of over $2000$. This drastic reduction suppresses the gradient signal, preventing the ``over-optimization'' that causes the model to drift from its pre-trained knowledge.

The Elastic Tether enables a self-regulating process that halts optimization once the policy has sufficiently diverged from the reference distribution, acting as an \emph{automatic, sample-wise early stopping mechanism}. By tethering the updated policy to the reference model and suppressing updates on well-learned samples, Reward-SFT preserves the pre-trained knowledge in $\pi_{\text{ref}}$.

\begin{wrapfigure}{r}{0.46\textwidth}
  \centering
  \vspace{-2pt} 
  \includegraphics[width=0.38\textwidth]{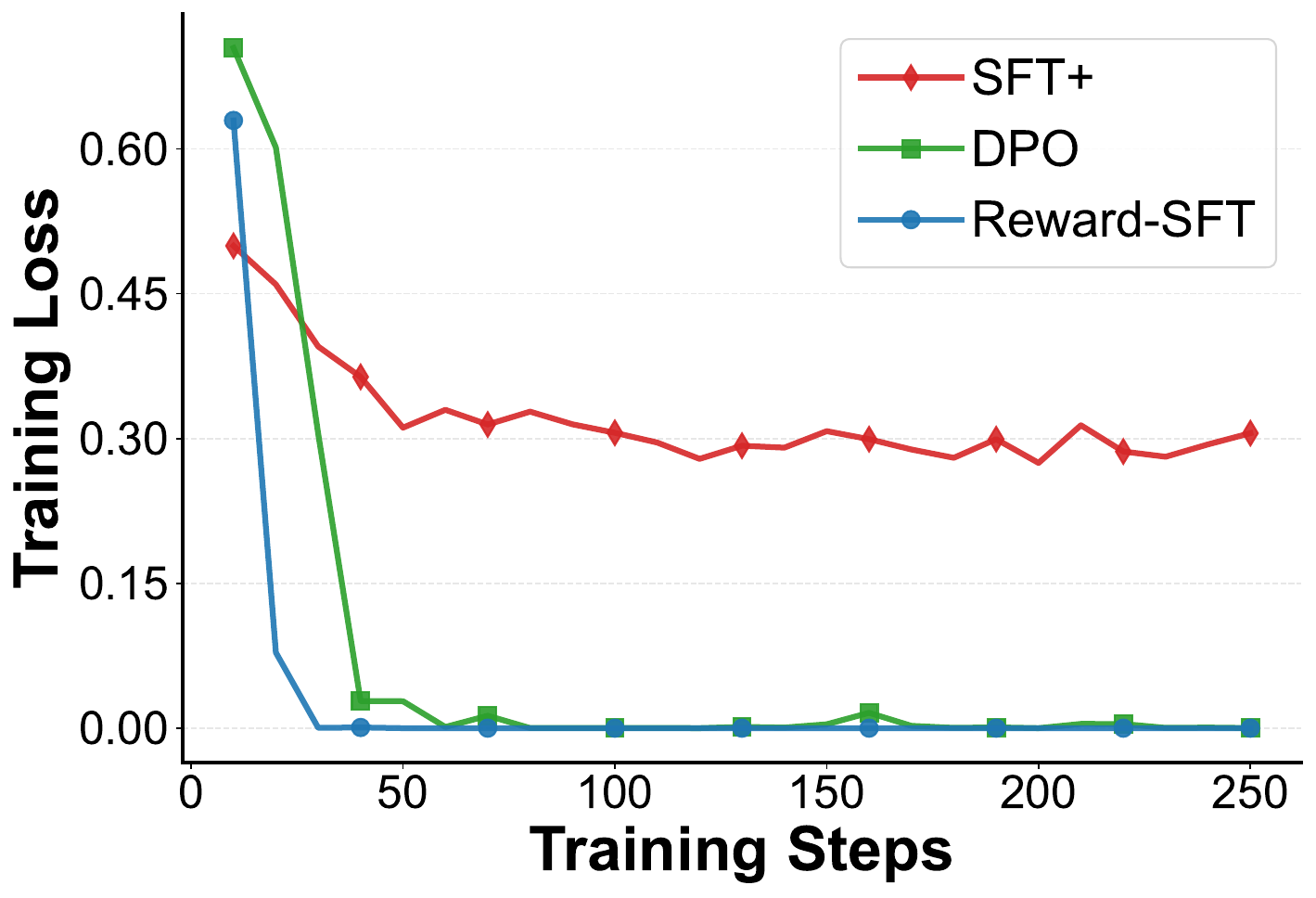}
  \caption{\textbf{Training loss curve.} Reward-SFT and DPO converge rapidly; SFT+ remains high due to absolute likelihood maximization.}
  \label{fig:loss}
  \vspace{-2pt} 
\end{wrapfigure}

These analytical insights are empirically substantiated by the training loss trajectories in \cref{fig:loss} and the reward evolution in \cref{fig:reward}. As shown in \cref{fig:loss}, the Reward-SFT loss rapidly converges to zero, a direct consequence of the definition in \cref{eq:rewardsft_loss}: as $r_\theta(x, y^+)$ increases, $\mathcal{L}_{\text{RW-SFT}}$ approaches zero. In contrast, the SFT objective drives continuous optimization, forcing superfluous parameter updates that degrade general capabilities. Furthermore, the reward trajectories in \cref{fig:reward} align well with our gradient analysis. As $r_\theta$ grows, the gradient of Reward-SFT vanishes. Consequently, the reward for chosen responses under Reward-SFT plateaus, whereas the SFT counterpart continues to rise without bound. In contrast to recent work suggesting that KL-divergence fails to mitigate forgetting \citep{chen2025retaining,shenfeld2025rl}, these analyses corroborate the critical role of KL-constrained reward in retaining knowledge during post-training.

\section{The Surgical Optimization Objective}
While preserving prior knowledge is essential, it is insufficient; the model must also achieve substantial in-domain reasoning improvements. This section uncovers the specific failure modes hindering this goal: the ``pull-up'' effect inherent in positive-only training (e.g., SFT, Reward-SFT) and the inadequacy of relative ranking (DPO) for reasoning with verifiable truth. To bridge this gap, we introduce a reward-based binary cross-entropy objective.

\subsection{The ``Pull-Up'' Effect}

\begin{figure}[t]
  \centering
  
  \includegraphics[width=0.38\textwidth]{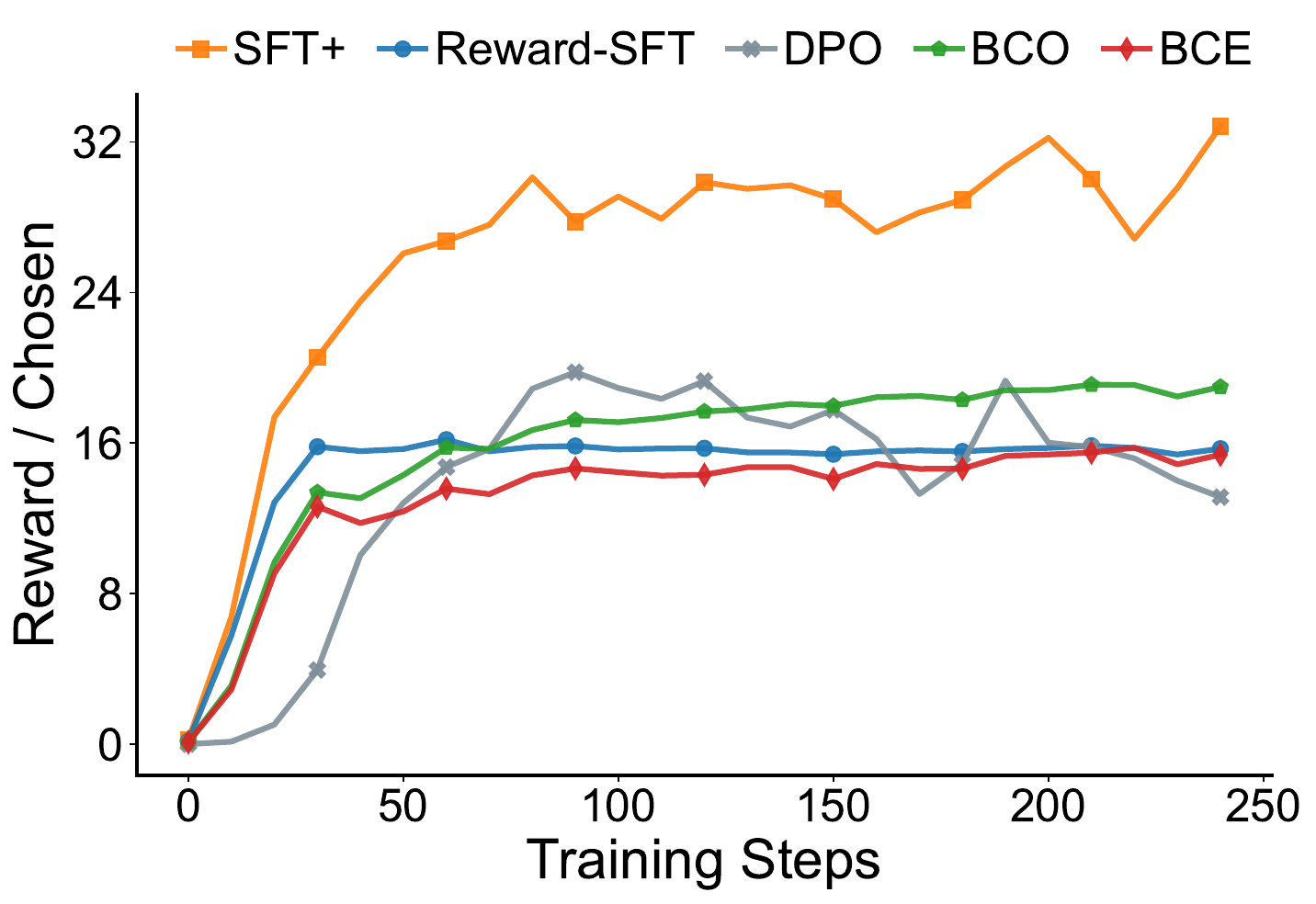}
  \hspace{1.8cm} 
  \includegraphics[width=0.38\textwidth]{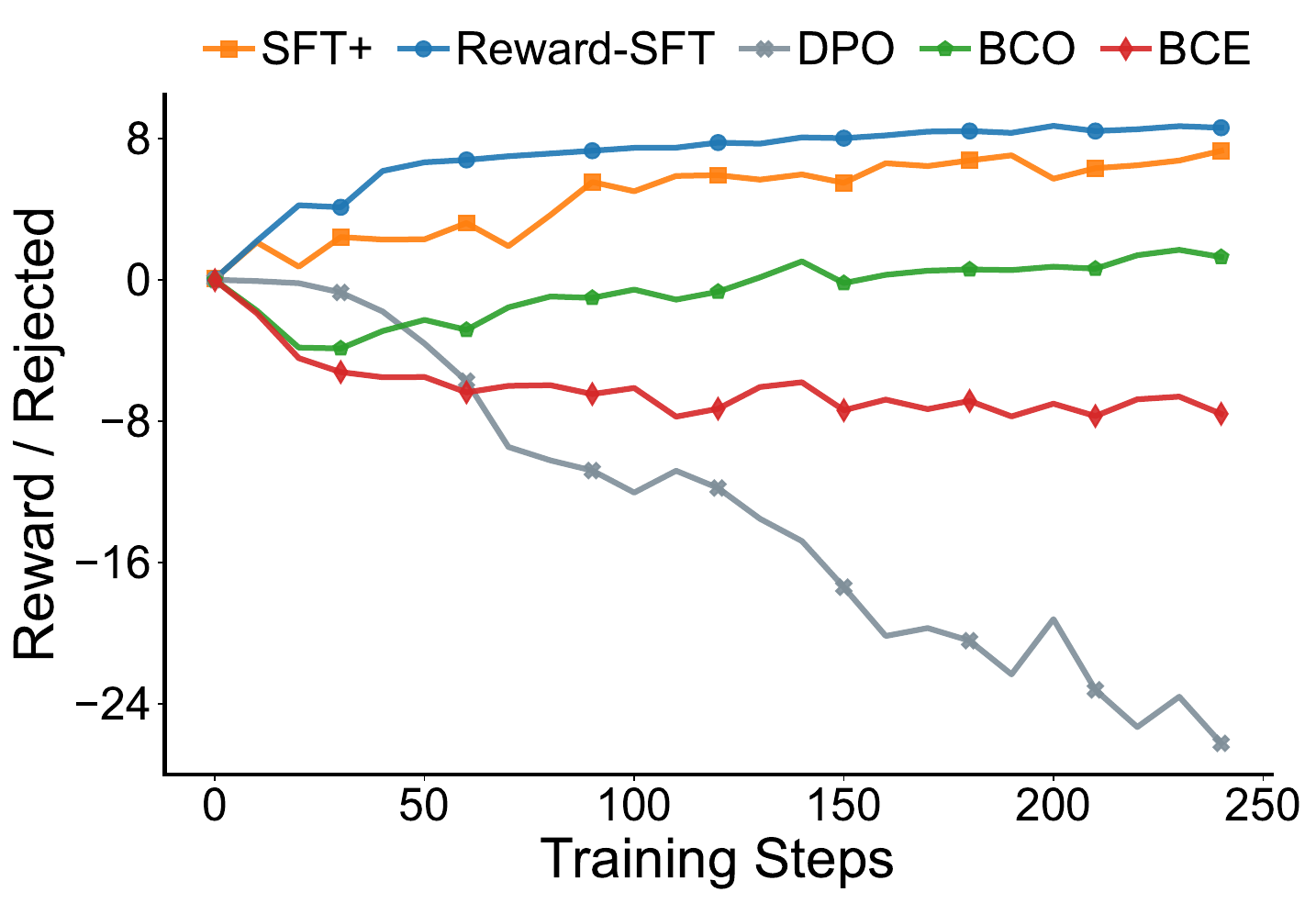}
  
  \caption{\textbf{Evolution of implicit rewards during training.} Left: reward scores for chosen responses; Right: reward scores for rejected responses. The reward measures how much the model's preference for the chosen response has grown relative to the initial state.}
  \label{fig:reward}
\end{figure}

While Reward-SFT effectively uses the ``Elastic Tether'' to prevent forgetting, it shares a key vulnerability with SFT when trained on our surgical rectification data. As shown in the right panel of \cref{fig:reward}, although we train only on chosen data $y^+$, the likelihood of rejected responses $y^-$ for both SFT+ and Reward-SFT \emph{inadvertently rises relative to the reference model}.

This effect is termed ``pull-up'' by \citet{renlearning}: under \textit{positive-only} supervision, the model indiscriminately raises the probability mass around the target, reinforcing similar but unwanted responses. In our rectification pipeline, the generated $y^+$ and $y^-$ share a vast majority of their token trajectories. Formally, let $y^\pm = p \oplus s^\pm$, where $p$ is the shared prefix and $s^\pm$ are the diverging suffixes. The gradient of the positive sample decomposes as $\nabla_\theta\log\pi_\theta(y^+|x) = \nabla_\theta\log\pi_\theta(p|x) + \nabla_\theta\log\pi_\theta(s^+|p,x)$. Because $\nabla_\theta\log\pi_\theta(p|x)$ is a shared prefix term with $\nabla_\theta\log\pi_\theta(y^-|x)$, maximizing the likelihood of $y^+$ inevitably raises the prefix component of $\log\pi_\theta(y^-|x)$. Consequently, the model fails to establish a sharp decision boundary at the critical point of divergence, underscoring the necessity of negative samples to suppress unwanted generation paths.
\subsection{The Insufficiency of Relative Ranking}
\label{sec:insufficiency}
Given the need for negative supervision, DPO appears to be a natural candidate. DPO models preference probabilities using the Bradley-Terry model \citep{bradley1952rank}, minimizing the negative log-likelihood of the margin $M = r_\theta(x, y^+) - r_\theta(x, y^-)$:
\begin{equation}
P(y^+ \succ y^-) = \sigma(r_\theta(x, y^+) - r_\theta(x, y^-))
\end{equation}
However, relative ranking is unsuitable for reasoning, where answers are objectively correct or incorrect, unlike subjective tasks (e.g., creative writing) where being ``better'' suffices. Reasoning requires maximizing the likelihood of the correct response $y^+$ while suppressing the incorrect $y^-$. Since gradient descent naturally follows the path of least resistance, the DPO loss can be minimized by decreasing the rejected reward $r(x,y^-)$ while leaving the chosen reward $r(x,y^+)$ stagnant---or even allowing it to drop, provided $r(x,y^-)$ drops faster. This phenomenon has been observed by \citet{rafailovr, pal2024smaug, renlearning}, and is corroborated by our results in \cref{fig:reward}: in later stages of DPO training, the reward for chosen responses fails to show sustained growth, while the reward for rejected ones decreases drastically. This suggests that DPO tends to optimize the margin by suppressing $y^-$ rather than reinforcing correct logic, rendering it insufficient for reasoning tasks.


\subsection{Reward-based Binary Cross Entropy Optimization}
\label{sec:method}

To address these limitations, we reframe reasoning optimization as a binary classification task. Unlike DPO's relative ranking between pairs, our approach decouples the supervision into independent positive and negative terms. We introduce two variants:

\textbf{SPoT-BCE.}\quad By decoupling the DPO objective in \cref{eq:dpo_loss}, we drive the model to independently maximize confidence in $y^+$ and minimize confidence for $y^-$. This results in the standard Binary Cross Entropy (BCE) loss:
\begin{equation}
    \mathcal{L}_{\text{SPoT-BCE}}\! =\! - \mathbb{E}_{\mathcal{D}} [ \log \sigma(r_\theta(x, y^+)) \!+\! \log \sigma(-r_\theta(x, y^-))]
    \label{eq:bce_loss}
\end{equation}
The implicit reward $r_\theta(x, y)$ here serves as the classification logit. This formulation provides a dense supervision signal while effectively retaining pre-trained knowledge.

\textbf{SPoT-BCO.}\quad While Binary Classifier Optimization (BCO) \citep{jung2025binary} was originally designed for unpaired data (e.g., thumbs-up/down only), we identify it as uniquely suitable for our paired reasoning data. BCO introduces a reward-shift $\delta$ to the BCE loss:
\begin{equation}
\begin{split}
    \mathcal{L}_{\text{SPoT-BCO}} = - \mathbb{E}_{x, y^+, y^- \sim \mathcal{D}} \big[ \log \sigma\left(r_\theta(x, y^+)\!-\!\delta\right) +
    \log \sigma\left(-\left(r_\theta(x, y^-)\!-\!\delta\right)\right)\big]
\end{split}
    \label{eq:bco_loss}
\end{equation}
where $\delta=\frac{1}{2}\cdot\mathbb{E}_{x, y^+, y^- \sim \mathcal{D}}[r_\theta(x, y^+)+r_\theta(x, y^-)]$. In practice, $\delta$ is computed as an exponential moving average of the batch statistics, and a stop-gradient is applied so that $\delta$ does not contribute to the backward pass (i.e., $\delta$ is treated as a constant when computing $\nabla_\theta \mathcal{L}_{\text{SPoT-BCO}}$).

Our contribution lies not in the BCE/BCO loss forms themselves, but in identifying their decoupled structure as uniquely suited to reasoning, where verifiable correctness demands independent reinforcement of $y^{+}$ and suppression of $y^{-}$; and the theoretical analysis of $\delta$ in the following subsection.

\subsection{Theoretical Analysis of \texorpdfstring{$\delta$}{delta}}
The difference between SPoT-BCE and SPoT-BCO lies in the reward shift $\delta$. While \citet{jung2025binary} frame $\delta$ as providing a tighter upper bound on DPO, we offer a different viewpoint and analyze its specific role in reasoning alignment.

The implicit reward $r_\theta(x,y)$ defined in \cref{eq:reward_def} is a simplified version of the theoretical optimal reward $r'(x,y)$. The relationship derived from the KL-constrained optimal policy $\pi^*$ is given by:
\begin{equation}
    r'(x,y) = \beta \log \frac{\pi^*(y|x)}{\pi_{\text{ref}}(y|x)} + \beta \log Z(x)
    \label{eq:theoretical_reward}
\end{equation}
where $Z(x)$ is the partition function. While DPO eliminates $\beta \log Z(x)$ via paired subtraction, BCO admits no analogous cancellation. Comparing \cref{eq:bco_loss} with \cref{eq:theoretical_reward} reveals that the shift term $\delta$ serves as a functional proxy for the intractable partition term $\beta \log Z(x)$.

\textbf{Mitigating Saturation for In-Domain Reasoning.}\quad As training progresses, the average implicit reward $r_\theta$ may increase (see \cref{fig:reward}). In SPoT-BCE ($\delta=0$), these elevated scores push the sigmoid input into saturation ($\sigma(\cdot) \to 1$), causing gradients to vanish. This saturation may halt optimization prematurely. By adapting $\delta$ via the exponential moving average, SPoT-BCO dynamically adjusts the threshold to enable further updates of the positive term $\log \sigma\left(r_\theta(x, y^+)\!-\!\delta\right)$, yielding higher rewards for chosen responses than SPoT-BCE (see \cref{fig:reward}) and, consequently, better performance on in-domain reasoning tasks.

\textbf{The Trade-off with OOD Generalization.}\quad Meanwhile, the early saturation in SPoT-BCE naturally limits the deviation of $\pi_\theta$ from $\pi_{\text{ref}}$. By relaxing this constraint, the adaptive shift in SPoT-BCO permits further optimization, leading to a larger KL divergence from the reference model. Consequently, SPoT-BCE retains prior knowledge more faithfully than SPoT-BCO.

\subsection{Surgical Post-Training}

\textit{Surgical Post-Training} takes its name from the synergistic design of our data pipeline and optimization objective. (1) Our data pipeline yields a dataset in which $y^-$ and $y^+$ share the majority of their token trajectories, diverging only at decision-critical parts. (2) For the tokens within the shared prefix of $y^+$ and $y^-$, the gradients for the positive and negative terms counteract each other, suppressing updates on these tokens. Consequently, parameter updates concentrate on the \textit{divergent} tokens, acting as a form of \textit{``surgery''} on the model---precisely correcting the reasoning failure while minimizing impact on the original distribution.

\section{Experiments}
\begin{table*}[t]
    \centering
        \caption{\textbf{Main results for in-domain reasoning, OOD reasoning, and general instruction following (OOD non-reasoning).} We compare \methodName (with BCO loss) against standard baselines: SFT (rejection-sampled Gemini 2.5 Pro responses), RFT (rejection-sampled self-generated responses), and SFT+ (rectified $y^+$). Direct SFT leads to performance degradation on both in-domain and OOD tasks. RFT fails to yield clear in-domain improvements and causes OOD regression. While SFT+ improves in-domain reasoning with near-distribution data, it still suffers OOD losses. In contrast, \methodName demonstrates superiority in both in-domain and OOD performance. We report avg@16 for AIME24, AIME25, and AMC23, and avg@5 for the others.}
    \label{tab:main_results}
    
    \resizebox{\textwidth}{!}{
        \begin{tabular}{l ccccccc ccc c l}
            \toprule
            \multirow{2}{*}{\textbf{Model}} & \multicolumn{7}{c}{\textbf{In-domain Reasoning}} & \multicolumn{3}{c}{\textbf{OOD Reasoning}} & \textbf{General} & \multirow{2}{*}{\textbf{Avg}} \\
            \cmidrule(lr){2-8} \cmidrule(lr){9-11} \cmidrule(lr){12-12}
             & AIME24 & AIME25 & AMC23 & Math500 & Minerva & Olympia & \textbf{Avg} & GPQA-D & Connect4 & \textbf{Avg} & IFEval & \\
            \midrule
            Qwen3-8b & 22.0 & 19.3 & 66.5 & 82.8 & 37.9 & 52.3 & 46.8 & 48.9 & 10.9 & 29.9 & 83.0 & 47.1 \\
            Qwen3-8b+SFT & 15.3 & 13.3 & 56.0 & 78.2 & 38.6 & 44.3 & 41.0 \negat{5.8} & 39.0 & 12.0 & 25.5 \negat{4.4} & 79.6 \negat{3.4} & 41.8 \negat{5.3} \\
            Qwen3-8b+RFT & 27.3 & 20.0 & 59.0 & 83.0 & 40.4 & 53.8 & 47.3 \pos{0.5} & 49.1 & 3.1 & 26.1 \negat{3.8} & 81.5 \negat{1.5} & 46.4 \negat{0.7} \\
            Qwen3-8b+\textbf{SFT+} & \textbf{29.3} & 24.7 & 67.3 & 84.8 & \textbf{42.5} & 54.5 & 50.5 \pos{3.7} & 47.3 & 14.1 & 30.7 \pos{0.8} & 80.0 \negat{3.0} & 49.4 \pos{2.3} \\
            Qwen3-8b\textbf{+\methodName} & 28.0 & \textbf{27.3} & \textbf{71.5} & \textbf{87.4} & 39.7 & \textbf{58.5} & \textbf{52.1} \pos{5.3} & 46.8 & \textbf{36.0} & \textbf{41.4} \pos{11.5} & \textbf{84.8} \pos{1.8} & \textbf{53.3} \pos{6.2} \\
            
            \midrule
            Llama3.1-8b-Instruct & 3.3 & \textbf{2.7} & 19.5 & 47.0 & 22.4 & \textbf{16.9} & 18.6 & 28.5 & 5.0 & 16.8 & \textbf{73.6} & 24.3 \\
            Llama3.1-8b-Instruct+SFT & 3.3 & 2.0 & 21.5 & 46.8 & 19.5 & 15.1 & 18.0 \negat{0.6} & 29.6 & 1.7 & 15.7 \negat{1.1} & 62.1 \negat{11.5} & 22.4 \negat{1.9} \\
            Llama3.1-8b-Instruct+RFT & 3.3 & 0.0 & 24.0 & 46.0 & 19.1 & 15.3 & 18.0 \negat{0.6} & \textbf{32.3} & 2.0 & 17.2 \pos{0.4} & 71.2 \negat{2.4} & 23.7 \negat{0.6} \\
            Llama3.1-8b-Instruct+\textbf{SFT+} & 3.3 & 0.7 & 26.0 & \textbf{48.6} & 25.7 & 15.3 & 19.9 \pos{1.3} & 31.9 & 1.5 & 16.7 \negat{0.1} & 68.6 \negat{5.0} & 24.6 \pos{0.3} \\
            Llama3.1-8b-Instruct\textbf{+\methodName} & \textbf{4.0} & 2.0 & \textbf{27.0} & \textbf{48.6} & \textbf{26.1} & 16.3 & \textbf{20.7} \pos{2.1} & 30.0 & \textbf{7.0} & \textbf{18.5} \pos{1.7} & 73.2 \negat{0.4} & \textbf{26.0} \pos{1.7} \\
            \bottomrule
        \end{tabular}
     }
\end{table*}

In this section, we present the experimental results of \methodName, followed by ablation studies that validate the effectiveness of our optimization objective (\cref{sec:exper_loss}) and our data pipeline (\cref{sec:exper_data}). Experiments are conducted on the English subset of DAPO-Math-17k \citep{yu2025dapo}. Full experimental settings are provided in Appendix \ref{appd:exp_setting}.

\subsection{Results}
We compare \methodName against: (1) \textbf{SFT:} supervised fine-tuning on rejection-sampled responses from Gemini 2.5 Pro. (2) \textbf{RFT:} rejection sampling fine-tuning using self-generated responses. (3) \textbf{SFT+:} supervised fine-tuning on the rectified positive samples $y^+$. The main results on Qwen3-8B and Llama-3.1-8B-Instruct are summarized in \cref{tab:main_results}.

\textbf{Catastrophic forgetting under SFT.}\quad SFT compromises general capabilities and OOD reasoning. This degradation is even more pronounced on Llama-3.1-8B-Instruct, where IFEval accuracy drops by 11.5 points. Furthermore, the drastic distributional shift between Gemini 2.5 Pro and the policy models erodes even in-domain reasoning (e.g., Qwen3-8B drops from 46.8\% to 41.0\%). The results confirm that unconstrained likelihood maximization on off-policy data shifts the model too aggressively, damaging its prior knowledge.

\textbf{Proximity alone is insufficient.}\quad Despite using ``surgical'' data proximal to the policy model, SFT+ improves in-domain reasoning yet fails to prevent forgetting on general tasks. Even with more ``on-policy'' data, RFT yields negligible in-domain improvements but incurs notable forgetting. These results indicate that data proximity alone cannot mitigate the forgetting induced by the SFT objective; without explicit regularization, the model's broad capabilities inevitably degrade.

\textbf{\methodName: superior reasoning with retained knowledge.}\quad \methodName outperforms all baselines on average across in-domain and OOD metrics on both Qwen3-8B and Llama-3.1-8B-Instruct. Moreover, \methodName even boosts general instruction following on Qwen3-8B. This demonstrates the effectiveness of both our surgical data rectification pipeline and our optimization objective. Additional general-capability results on TruthfulQA and MMLU-Pro are reported in Appendix~\ref{appd:general_caps}.

\subsection{Comparison on Loss Objectives} 
\label{sec:exper_loss}
\begin{table*}[t]
    \centering
    \caption{\textbf{Comparison on loss objectives.} We find that SPoT-BCO yields the best overall results. SPoT-BCE is more competitive on keeping general abilities. DPO mitigates catastrophic forgetting but shows no clear improvements for in-domain reasoning. Although DFT~\cite{wu2025generalization} is designed for generalization, we observe a clear degradation on OOD tasks. }
    \label{tab:loss}
    \resizebox{\textwidth}{!}{
        \begin{tabular}{l ccccccc ccc c l}
            \toprule
            \multirow{2}{*}{\textbf{Model}} & \multicolumn{7}{c}{\textbf{In-domain Reasoning}} & \multicolumn{3}{c}{\textbf{OOD Reasoning}} & \textbf{General} & \multirow{2}{*}{\textbf{Avg}} \\
            \cmidrule(lr){2-8} \cmidrule(lr){9-11} \cmidrule(lr){12-12}
             & AIME24 & AIME25 & AMC23 & Math500 & Minerva & Olympia & \textbf{Avg} & GPQA-D & Connect4 & \textbf{Avg} & IFEval & \\
            \midrule            
            \textbf{SPoT-BCO} & 28.0 & \textbf{27.3} & \textbf{71.5} & 87.4 & 39.7 & \textbf{58.5} & \textbf{52.1}\pos{5.3}  & 46.8 & \textbf{36.0} & 41.4\pos{11.5} & 84.8\pos{1.8} & \textbf{53.3}\pos{6.2}  \\
            
            \textbf{SPoT-BCE} & \textbf{33.3} & 25.3 & 67.0 & \textbf{87.6} & 40.1 & 55.4 & 51.5\pos{4.7} & \textbf{49.5} & 31.1 & 40.3\pos{10.4} & \textbf{85.8}\pos{2.8} & 52.8\pos{5.7} \\
 
            DPO & 22.0 & 19.3 & 62.5 & 85.2 & \textbf{40.8} & \underline{51.6} & 46.9\pos{0.1} & 48.9 & 31.8 & 40.4\pos{10.5} & 84.7\pos{1.7} & 49.6\pos{2.5} \\
            
            Reward-SFT & \underline{20.0} & \underline{16.7} & \underline{60.5} & 83.0 & \underline{39.0} & 52.3 & \underline{45.3}\negat{1.5} & 49.2 & 35.2 & \textbf{42.2}\pos{12.3} & 83.7\pos{0.7} & 48.9\pos{1.8} \\

            DFT & 24.0 & 23.3 & 70.5 & \underline{82.4} & 40.4 & 52.4 & 48.8\pos{2.0} & \underline{45.6} & \underline{3.7} & \underline{24.7}\negat{5.2} & \underline{79.5}\negat{3.5} & \underline{46.9}\negat{0.2} \\
            \bottomrule
        \end{tabular}
    }
\end{table*}

To validate our choice of optimization objective, we compare SPoT-BCO and SPoT-BCE against DPO, Reward-SFT, and DFT~\citep{wu2025generalization} on Qwen3-8B. The results are shown in \cref{tab:loss}. SPoT-BCO yields the best overall average and in-domain reasoning accuracy. The adaptive boundary $\delta$ in BCO creates a tighter optimization constraint, pushing the model toward clearer decision boundaries on complex math problems. SPoT-BCE serves as a strong alternative, better preserving general capabilities due to its strict regularization and achieving the highest scores on IFEval (85.8\%) and GPQA-D (49.5\%). In contrast, while DPO mitigates catastrophic forgetting (IFEval 84.7\%), it fails to improve in-domain reasoning. This suggests that relative ranking is insufficient for learning correct reasoning paths. Reward-SFT resists forgetting thanks to regularization, yet its in-domain reasoning declines due to the ``pull-up'' effect. Finally, despite being designed for generalization, DFT exhibits clear degradation against the raw SFT objective in our experiments.
We further compare against the DPO+SFT recipe~\citep{wang2024enhancing} in Appendix~\ref{appd:dpo_sft}.

\subsection{The Effect of the Surgical Rectification Pipeline}
\label{sec:exper_data}

\begin{table*}[t]
    \centering
            \caption{\textbf{Ablation study on different data.} ``Rectified data'' refers to the dataset curated using our surgical rectification pipeline, while ``direct data'' uses the rejected-sampling direct answers from Gemini 2.5 Pro. $\gamma$ defines the threshold of change ratio in \Cref{eq:change_ratio}.}
    \label{tab:ablation}
    \resizebox{\textwidth}{!}{
        \begin{tabular}{l ccccccc ccc c l}
            \toprule
            \multirow{2}{*}{\textbf{Model}} & \multicolumn{7}{c}{\textbf{In-domain Reasoning}} & \multicolumn{3}{c}{\textbf{OOD Reasoning}} & \textbf{General} & \multirow{2}{*}{\textbf{Avg}} \\
            \cmidrule(lr){2-8} \cmidrule(lr){9-11} \cmidrule(lr){12-12}
             & AIME24 & AIME25 & AMC23 & Math500 & Minerva & Olympia & \textbf{Avg} & GPQA-D & Connect4 & \textbf{Avg} & IFEval & \\
            \midrule
            2k direct data & \underline{20.0} & \underline{14.7} & \underline{63.0} & \underline{81.6} & \underline{37.5} & \underline{47.6} & \underline{44.1} \negat{2.7} & \underline{42.6} & \underline{17.2} & 29.9 \negat{0.0}  & \underline{82.6} \negat{0.4} & \underline{45.2} \negat{1.9} \\
            2k rectified data, $\gamma=0.6$ & 26.0 & 19.3 & 67.0 & \textbf{87.8} & 41.2 & 52.4 & 49.0 \pos{2.2} & \textbf{49.8} & 26.0 & 37.9 \pos{8.0} & 84.1 \pos{1.1} & 50.4 \pos{3.3} \\
            4k rectified data, $\gamma=1$ & 27.3 & 22.7 & 64.0 & 87.6 & \textbf{41.5} & 58.2 & 50.2 \pos{3.4} & 49.2 & 33.3 & 41.3 \pos{11.4} & 83.6 \pos{0.6} & 51.9 \pos{4.8} \\
            4k rectified data, $\gamma=0.6$ & \textbf{28.0} & \textbf{27.3} & \textbf{71.5} & 87.4 & 39.7 & \textbf{58.5} & \textbf{52.1} \pos{5.3} & 46.8 & \textbf{36.0} & \textbf{41.4} \pos{11.5} & \textbf{84.8} \pos{1.8} & \textbf{53.3} \pos{6.2} \\
            \bottomrule
        \end{tabular}
    }
\end{table*}

We validate the impact of \textit{data proximity} by ablating two key components: the data source and the proximity constraint $\gamma$ in \cref{sec:data}. The results are summarized in \cref{tab:ablation}. First, training on ``rectified data'' significantly outperforms ``direct data'' (+5.2\%). Although SPoT-BCO already mitigates catastrophic forgetting on out-of-distribution data (only a 0.4-point IFEval drop), the substantial reasoning gain from rectified data confirms that proximal data is inherently more effective. Second, scaling the dataset size from 2k to 4k yields consistent improvements. Finally, when controlling for the final training set size, the subset filtered with $\gamma=0.6$ achieves the highest average score (53.2\%), surpassing the unfiltered configuration ($\gamma=1$). This further confirms that enforcing data proximity is critical for maximizing performance.

\subsection{Superior Initialization for Reinforcement Learning}
\label{sec:exper_rl}

\begin{table*}[t]
    \centering
    \caption{\textbf{Comparison with GRPO.} \methodName + GRPO: GRPO initialized from \methodName checkpoint.}
    \label{tab:rl}
    \resizebox{0.75\textwidth}{!}{
    \begin{tabular}{lcccccc}
\toprule
Method      & AIME 2024 & AIME 2025 & AMC 2023 & Connect4 & IFEval & \textbf{Avg} \\
\midrule
\methodName        & 28.0 \pos{6.0}  & 27.3 \pos{8.0}  & 71.5 \pos{5.0}  & 36.0 \pos{25.1} & 84.8 \pos{1.8}  & 49.5 \pos{9.2} \\
GRPO               & 36.3 \pos{14.3} & 28.7 \pos{9.4}  & 77.5 \pos{11.0} & 7.7 \negat{3.2}  & 81.2 \negat{1.8} & 46.3 \pos{6.0} \\
\methodName + GRPO & 41.3 \pos{19.3} & 36.7 \pos{17.4} & 86.0 \pos{19.5} & 29.4 \pos{18.5} & 82.1 \negat{0.9} & 55.1 \pos{14.8} \\
\bottomrule
\end{tabular}
}
\end{table*} 

To understand how \methodName affects subsequent RL, we evaluate GRPO training initialized from the \methodName checkpoint (\cref{tab:rl}). We observe that GRPO yields more in-domain improvements compared to \methodName, but induces some forgetting, evidenced by performance degradation on Connect4 (our contamination-free OOD benchmark) and a 1.8\% absolute decrease on IFEval. More importantly, \methodName provides a superior initialization for subsequent RL, significantly elevating the performance ceiling. \methodName + GRPO outperforms GRPO alone across all benchmarks (+7.2\% in-domain and +21.7\% OOD reasoning), alongside better general instruction-following capabilities. The results demonstrate that \methodName's knowledge injection establishes a stronger foundation that facilitates subsequent RL exploration.

\section{Related Work}
Recent studies show that RL mitigates catastrophic forgetting better than SFT by leveraging on-policy data \citep{shenfeld2025rl, chen2025retaining}. Complementary to this, we highlight the critical role of the KL-constrained reward formulation in DPO for knowledge preservation. However, DPO remains sub-optimal for reasoning tasks~\citep{renlearning,azar2024general}. Subsequent works improve DPO via synthetic data refinement \citep{pal2024smaug}, reference-free objectives~\citep{hong2024orpo,meng2024simpo}, iterative variants \citep{chen2024self,pang2024iterative}, and step-level supervision \citep{lai2024step}. In contrast, \methodName resolves these issues by introducing a rectification pipeline that produces contrasting pairs and by employing a binary loss in place of the rank-based objective in DPO. Furthermore, unlike on-policy distillation methods, which require logit-level supervision \citep{agarwal2024gkd, lu2025onpolicydistillation} unavailable from proprietary LLMs, \methodName operates with query-based API access alone, requiring no teacher probabilities. A more comprehensive review, including positioning against MiniLLM, GKD, OPSD, GAD, and KD-RL, is provided in Appendix~\ref{appd:related_work}.

\section{Conclusion and Discussion}
We present \methodName, a proximal on-policy distillation framework that combines a \textit{surgical rectification pipeline} with a \textit{binary optimization objective} to efficiently inject reasoning knowledge without catastrophic forgetting. The pipeline produces minimal-edit on-policy supervision by correcting only erroneous steps, preserving the model's prior token distribution. Theoretically and empirically, we show that the KL-constrained reward formulation acts as a sample-wise early stopping mechanism that retains knowledge, and that a \textit{binary} objective is critical for rigid reasoning tasks where verifiable truth makes pairwise preferences ill-suited. With only 4k rectified pairs and minutes of training, \methodName delivers consistent gains across in-domain and OOD benchmarks, and substantially raises the performance ceiling of subsequent GRPO, positioning proximal on-policy distillation as a lightweight bridge between SFT and RL.

\textbf{Limitations and future work.}\quad Due to resource constraints, our experiments focus on smaller models (up to 8B) and text-based mathematical reasoning (additional Qwen3-1.7B results in Appendix~\ref{appd:scale_general}). Performance also depends on the Oracle's capability; we report a sensitivity analysis in Appendix~\ref{appd:weak_oracle}, leaving a larger-scale study across diverse Oracles for future work. Promising directions include scaling to larger models and extending \methodName to multimodal, code, and agentic domains, where surgical rectification of long, structured trajectories may be especially valuable.
\section*{Acknowledgements}
This work is supported by Hong Kong Research Grant Council - General Research Fund (Grant No. 17211024) and HKU Seed Fund for PI Research.

\bibliographystyle{icml2026}
\setlength{\bibsep}{1.5pt plus 0.3ex}
\bibliography{example_paper}

\newpage

\appendix
\section{Extended Related Work}\label{appd:related_work}
\textbf{Catastrophic Forgetting.}\quad Recent work highlights a divergence between SFT and RL: while SFT is prone to overfitting and catastrophic forgetting, RL demonstrates superior generalization \citep{chu2025sft,huan2025does}. To explain this, \citet{mukherjee2025reinforcement} observed that RL updates are typically sparse, affecting only specific parameter subnetworks, whereas SFT triggers dense, global updates. Alternatively, \citet{shenfeld2025rl} and \citet{chen2025retaining} argued that on-policy data generation---rather than the optimization objective---is the primary driver of knowledge retention. However, our analysis in \cref{sec:reward} suggests that the optimization objective remains critical: specifically, the implicit regularization (KL-constraint) inherent in reward-based objectives is essential for preserving prior knowledge.

\textbf{DPO for Reasoning.}\quad DPO tends to overly penalize rejected responses rather than strengthening the chosen ones~\citep{renlearning,azar2024general,pal2024smaug}, making it sub-optimal for reasoning. To mitigate this, several variants have emerged: \citet{azar2024general} introduced IPO to regularize against overfitting; \citet{pal2024smaug} and \citet{meng2024simpo} encouraged a larger margin to incentivize learning from positive examples; and others \citep{hong2024orpo,wang2024enhancing} integrated SFT objectives to reinforce preferred responses. 

Furthermore, methods such as KTO \citep{ethayarajh2024kto} and BCO \citep{jung2025binary} use unpaired or binary feedback to reduce annotation costs. While BCO was originally designed for data efficiency, we show, both analytically and empirically, that its loss structure is uniquely suited to address the ranking inefficiencies of DPO while mitigating catastrophic forgetting during post-training for reasoning.

Beyond general alignment, other works also leverage DPO to enhance reasoning capabilities. \citet{chen2024self} and \citet{pang2024iterative} employed iterative self-improvement, distinguishing model generations from human data. \citet{lai2024step} introduced Step-DPO, which includes a data rectification pipeline for precise step-level feedback. \methodName distinguishes itself from Step-DPO in two critical aspects. First, while Step-DPO relies on self-correction---limited by the model's intrinsic capabilities---we employ an Oracle-guided correction pipeline. Second, whereas Step-DPO retains the standard DPO objective, \methodName uses a reward-based binary cross-entropy objective. This approach more effectively addresses the sub-optimality of DPO for reasoning while mitigating forgetting in the post-training phase.

\textbf{On-Policy Distillation.}\quad \methodName can be viewed as a form of on-policy distillation (OPD), in which the student is trained to recover from its own self-generated mistakes under teacher supervision~\citep{agarwal2024gkd,gu2024minillm}. Recent analyses \citep{li2026rethinking,song2026opdsurvey} and large-scale practice \citep{yang2025qwen3,lu2025onpolicydistillation} confirm that OPD delivers substantial reasoning enhancements with high computational efficiency, validating the on-policy-with-dense-supervision recipe. However, these approaches rely on logit-level supervision from the teacher, which is unavailable for the strongest proprietary models accessible only via inference APIs. Existing black-box workarounds either fall back to off-policy \textsc{SeqKD} \citep{kim2016seqkd} on teacher-generated text, or, more recently, replace the missing logit signal with an adversarially trained on-policy reward model \citep{ye2025gad}. Orthogonally, OPSD \citep{zhao2026opsd} dispenses with an external teacher altogether by letting a single model play both roles under different conditioning contexts.

\methodName sits in this design space at a deliberately different point.
First, it operates strictly under \emph{query-based API access}---requiring neither teacher logits nor a learned discriminator---which
makes it directly compatible with frontier proprietary teachers and
sidesteps the well-known instability of adversarial training in
\citet{ye2025gad}. Second, in place of the standard divergence-matching
loss, \methodName adopts a \emph{regularized binary cross-entropy}
objective explicitly designed to mitigate catastrophic forgetting; this avoids the RL/KL trade-off that \citet{xu2025kdrl} report as hyperparameter-sensitive, and, unlike \citet{zhao2026opsd}, admits genuinely stronger external teachers rather than constraining the student to its own conditional distribution.

\clearpage

\section{Rectification Prompt}
\label{appd:prompt}
\begin{tcolorbox}[colback=green!3, title=Rectification Prompt Without Ground Truth, enhanced, sharp corners, fonttitle=\small]
\begin{Verbatim}[breaklines=true, formatcom=\relax, breakanywhere=true, breaksymbolleft=, breaksymbolright=,fontsize=\scriptsize]
Act as a helpful teaching assistant. Your goal is to revise a student model's answer to make it correct, while maintaining the student model's original writing style, tone, and formatting. The final result should look as if the student model had solved the problem correctly on its first try.

You should first solve the problem independently and do the following:
1. Identify the correct parts of the student model's answer and keep them.
2. Replace the incorrect parts with correct reasoning.
3. Carefully match the student model's original writing style, including their tone, vocabulary, formatting and sentence structure.

**IMPORTANT OUTPUT FORMAT:**
1. First output ``=== CORRECTED STARTED ==='' followed by the corrected answer
2. Ends with the corrected answer in the format: 'Therefore, the final answer is: $\\boxed{{ANSWER}}$.'
3. Then output ``=== CORRECTED ENDED ==='' at the end of the corrected trace
4. Do not output meta-phrases like "Here is the corrected version"
\end{Verbatim}
\end{tcolorbox}

\begin{tcolorbox}[colback=green!3, title=Rectification Prompt With Ground Truth, enhanced, sharp corners, fonttitle=\small]
\begin{Verbatim}[breaklines=true, formatcom=\relax, breakanywhere=true, breaksymbolleft=, breaksymbolright=,fontsize=\scriptsize]
Act as a helpful teaching assistant. Your goal is to revise a student model's answer to make it correct, while maintaining the student model's original writing style, tone, and formatting. The final result should look as if the student model had solved the problem correctly on its first try.

You should compare the student model's answer with the Reference Ground Truth and do the following:
1. Identify the correct parts of the student model's answer and keep them.
2. Replace the incorrect parts with correct reasoning.
3. Carefully match the student model's original writing style, including their tone, vocabulary, and sentence structure.

**IMPORTANT OUTPUT FORMAT:**
1. First output ``=== CORRECTED STARTED ==='' followed by the corrected answer
2. Ends with the corrected answer in the format: 'Therefore, the final answer is: $\\boxed{{ANSWER}}$.'
3. Then output ``=== CORRECTED ENDED ==='' at the end of the corrected trace
4. Do not output meta-phrases like "Here is the corrected version"
\end{Verbatim}
\end{tcolorbox}

\section{Rectification Samples}
\begin{figure}[htbp]
\centering
\includegraphics[width=1\linewidth]{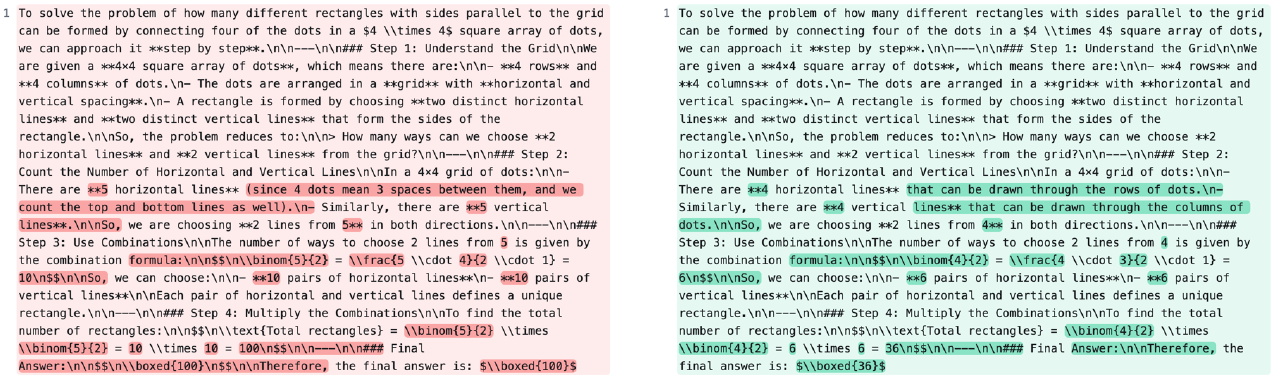}
\caption{\textbf{Random Rectification Example 1.} Left: answer from Qwen3-8B; Right: rectification by Gemini 2.5 Pro.}
\end{figure}
\begin{figure}[htbp]
\centering
\includegraphics[width=1\linewidth]{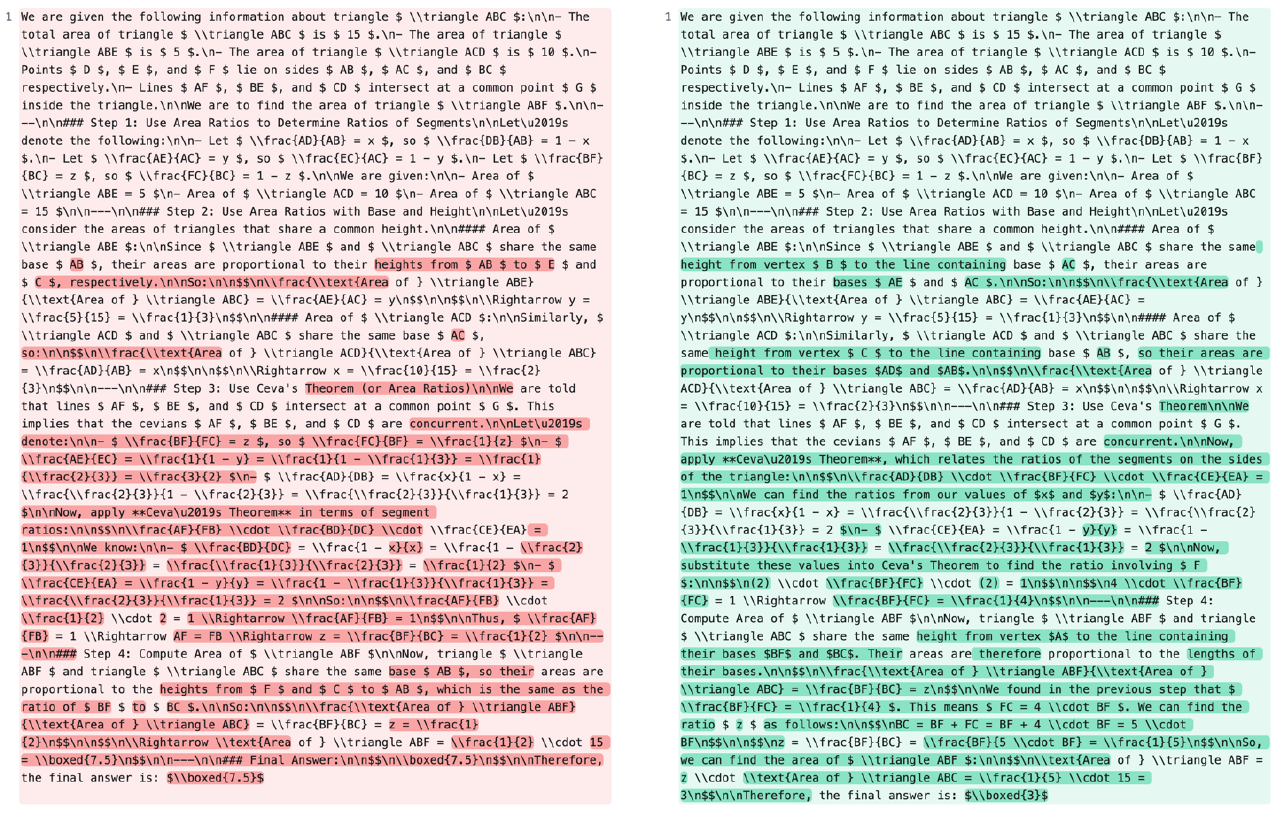}
\caption{\textbf{Random Rectification Example 2.} Left: answer from Qwen3-8B; Right: rectification by Gemini 2.5 Pro.}
\end{figure}
\begin{figure}[htbp]
\centering
\includegraphics[width=1\linewidth]{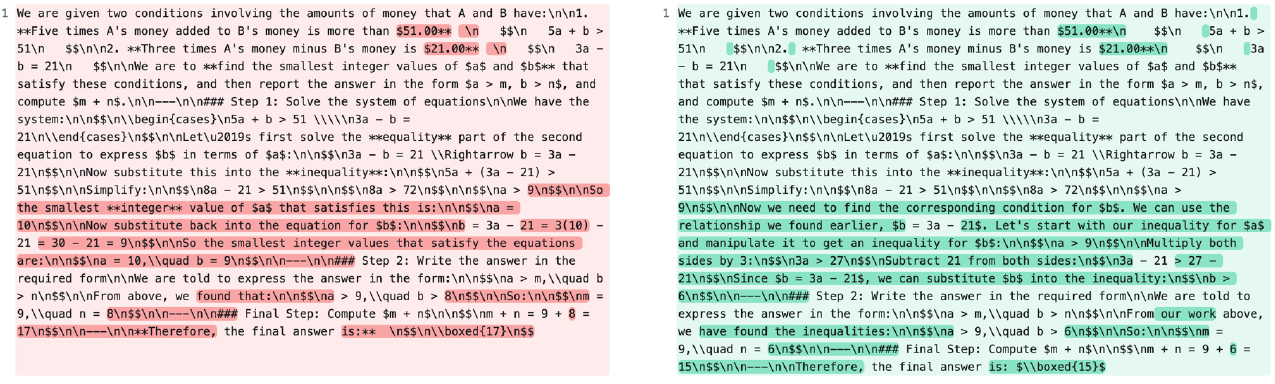}
\caption{\textbf{Random Rectification Example 3.} Left: answer from Qwen3-8B; Right: rectification by Gemini 2.5 Pro.}
\end{figure}

\begin{figure}[htbp]
\centering
\includegraphics[width=1\linewidth]{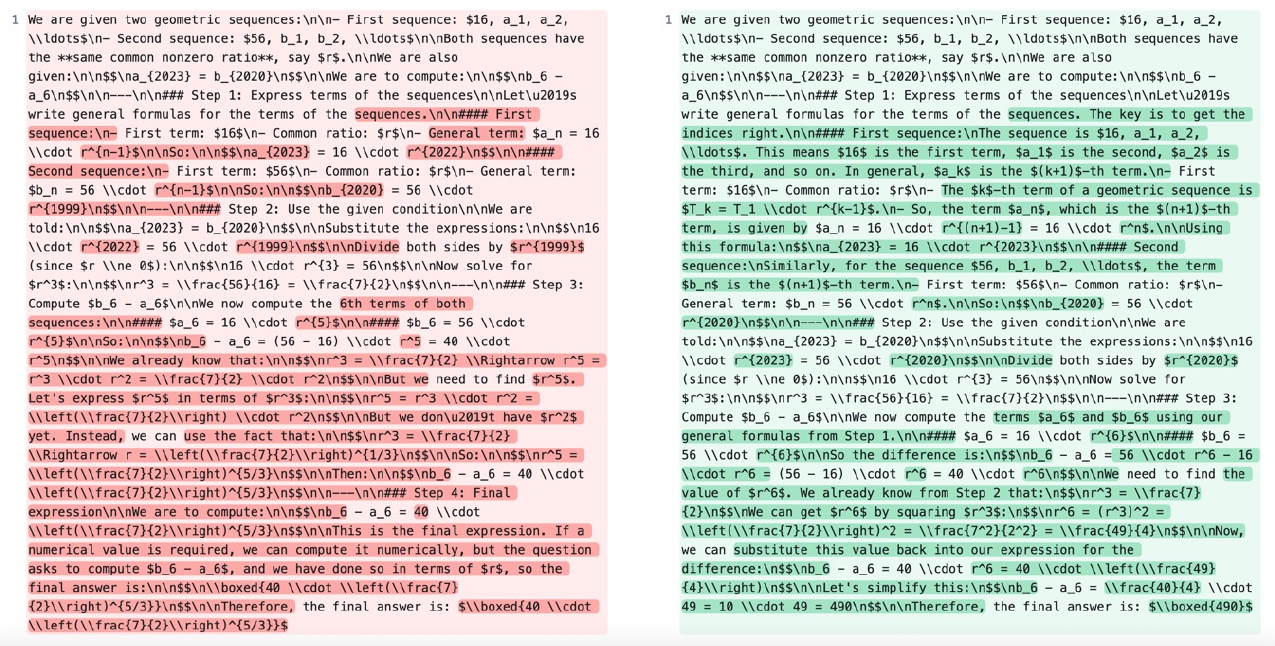}
\caption{\textbf{Random Rectification Example 4.} Left: answer from Qwen3-8B; Right: rectification by Gemini 2.5 Pro.}
\end{figure}

\clearpage

\section{Sensitivity to Oracle Capability}
\label{appd:weak_oracle}
We replace the default Oracle (Gemini~2.5~Pro) with the weaker open-source \texttt{Qwen3-30B-A3B-2507}; \methodName remains effective (\cref{tab:weak_oracle}).

\begin{table}[ht]
\centering
\caption{\methodName with a weaker open-source Oracle (\texttt{Qwen3-30B-A3B-2507}) on Qwen3-8B. Even without a frontier API, the framework yields consistent gains.}\label{tab:weak_oracle}
\small
\begin{tabular}{lccccc}
\toprule
Model & AIME24 & AIME25 & AMC23 & Connect4 & IFEval \\
\midrule
Base                                       & 22.0 & 19.3 & 66.5 & 10.9 & 83.0 \\
\methodName (Qwen3-30B-A3B Oracle)     & 25.6 & 22.5 & 67.0 & 26.9 & 82.9 \\
\bottomrule
\end{tabular}
\end{table}

\section{Prompt Sensitivity of the Oracle}
\label{appd:prompt_sensitivity}

To verify that the gains are not artifacts of careful prompt engineering, we replace our default rectification prompt with a deliberately simplified variant:

\begin{quote}
\itshape
``Revise a student model's answer to make it correct, while maintaining the student model's original writing style, tone, and formatting. Identify the correct parts and keep them, and replace the incorrect parts with correct reasoning. Only output the corrected answer.''
\end{quote}

We re-curate 2k pairs with the simplified prompt and re-train under identical hyperparameters. The two prompts produce nearly identical performance across all benchmarks (\cref{tab:prompt_sensitivity}), indicating that the framework's effectiveness is not dependent on a specific prompt formulation.

\begin{table}[ht]
\centering
\caption{Prompt sensitivity of the Oracle on Qwen3-8B (2k pairs). Performance is robust to prompt variation.}\label{tab:prompt_sensitivity}
\small
\begin{tabular}{lccccc}
\toprule
Prompt & AIME24 & AIME25 & AMC23 & Connect4 & IFEval \\
\midrule
Default     & 26.0 & 19.3 & 67.0 & 26.0 & 84.1 \\
Simplified  & 26.5 & 18.7 & 66.5 & 27.0 & 84.1 \\
\bottomrule
\end{tabular}
\end{table}

\section{Rationale for the LCS Filtering Threshold}
\label{appd:lcs_threshold}

The LCS-based filter governs an intrinsic trade-off between curation efficiency and the on-policy constraint:
\begin{itemize}
    \item A \emph{strict} threshold (e.g., $0.5$) yields more on-policy data but discards a large fraction of candidate pairs, making valid pairs more expensive to gather.
    \item A \emph{loose} threshold (e.g., $0.7$) retains more pairs at the risk of admitting off-policy data.
\end{itemize}
We empirically find $0.6$ to be the sweet spot between curation efficiency and on-policy fidelity. Starting from 6{,}171 error samples generated by Qwen3-8B, we ablate the filtering threshold and report results in \cref{tab:lcs_threshold}.

\begin{table}[ht]
\centering
\caption{Effect of the LCS filtering threshold on Qwen3-8B. The threshold $0.6$ used in the main paper balances the trade-off.}\label{tab:lcs_threshold}
\small
\begin{tabular}{lccccc}
\toprule
Setting & AIME24 & AIME25 & AMC23 & Connect4 & IFEval \\
\midrule
Base                                & 22.0 & 19.3 & 66.5 & 10.9 & 83.0 \\
\methodName, $\tau{=}0.5$ (1{,}998 pairs) & 19.3 & 22.0 & 64.0 & 23.7 & 84.5 \\
\methodName, $\tau{=}0.7$ (3{,}646 pairs) & 26.0 & 22.0 & 67.0 & 27.9 & 82.3 \\
\bottomrule
\end{tabular}
\end{table}

\textbf{Why a surface-level proxy suffices.}\quad A more rigorous measure of distributional proximity between rectified text and the student's own distribution would be the student's perplexity (PPL) on the rectified sequence. However, computing PPL requires a full forward pass per candidate. We thus adopt LCS as a cheap surface-level proxy, with the intuition that sequences whose tokens closely follow the student's sampling trajectory should also exhibit lower PPL under the student. To substantiate this, we computed the actual PPL on randomly sampled rectified pairs, partitioned by their LCS-based difference ratio:
\begin{itemize}
    \item 1k pairs with change ratio $<0.6$ (more on-policy by LCS): mean PPL $= 1.60$.
    \item 1k pairs with change ratio $>0.6$ (less on-policy by LCS): mean PPL $= 1.95$.
\end{itemize}
The relationship confirms that LCS is an effective proxy for student-side perplexity at curation time.

\section{Change Ratio of Rectification}
\label{appd:change_ratio}
\begin{figure}[htbp]
  \centering
  \includegraphics[width=0.43\linewidth]{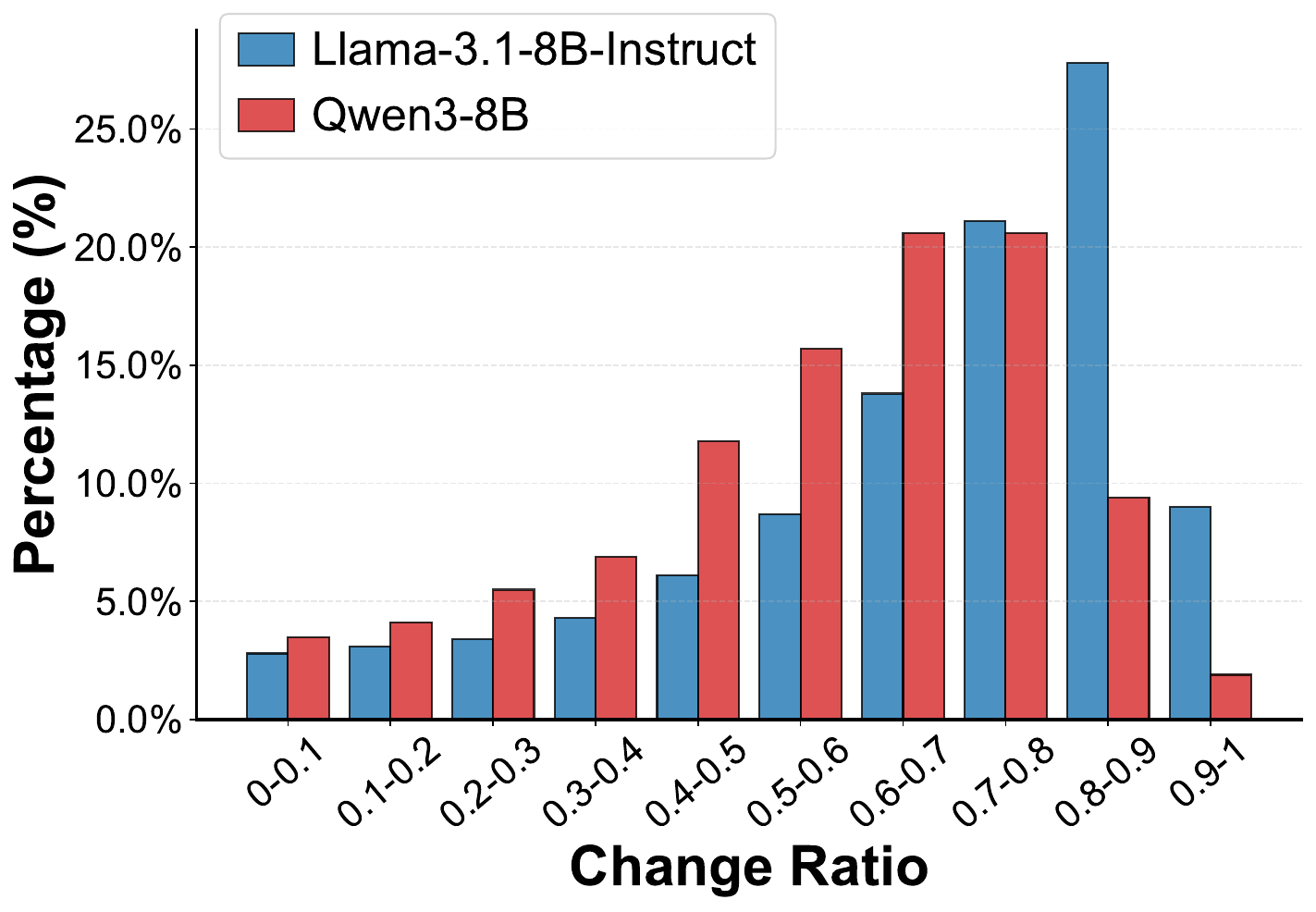}
  \caption{\textbf{Distributions of change ratio.} A higher change ratio indicates that reasoning failures occur earlier in the reasoning chain. The distribution shape is determined by the model ability and the data difficulty together.}\label{fig:sim_score}
  \vspace{-5pt}
\end{figure}

We visualize the distributions of change ratio, which measures the proportion of the reasoning chain that was modified during rectification. We calculate this ratio based on the edit distance between the original and rectified responses. The results are shown in \cref{fig:sim_score}.

\section{Cost and Efficiency Analysis}
\label{appd:cost}

\textbf{Oracle API cost.}\quad Curating the full 4k rectified pairs with Gemini~2.5~Pro (priced at \$10/1M output tokens) cost approximately \$278 in total and completed in roughly 40 minutes when run in parallel with student self-sampling. To eliminate reproducibility hurdles, we will release the full pipeline, including all curated training data, the curation and training codebases, model checkpoints, and evaluation scripts.

\textbf{Wall-clock and GPU-hour comparison with GRPO.}\quad We compare end-to-end training cost between GRPO and \methodName on Qwen3-8B under identical hardware ($8\times$ H800 80GB) in \cref{tab:time_cost}. \methodName is efficient, as it requires only one rollout per sample.

\begin{table}[ht]
\centering
\caption{Wall-clock and GPU-hour for GRPO and \methodName on Qwen3-8B. Teacher rectification is parallel with student sampling and therefore does not appear as a separate term.}\label{tab:time_cost}
\small
\begin{tabular}{lcc}
\toprule
Component & GRPO & \methodName \\
\midrule
Total time         & $\sim$13 h  & $\sim$1 h \\
Sampling                  & 5.56 h      & 43 min \\
Policy training            & 5.12 h      & 16 min \\
Reference log-prob        & 2.25 h      & (in training) \\
Teacher rectification       & ---         & overlapped with sampling \\
\midrule
Total GPU hours             & $\sim$104 & $\sim$8 \\
\bottomrule
\end{tabular}
\end{table}

\section{Detailed Results with Standard Deviations}
\label{sec:detailed_results_std}

To ensure the statistical significance of our experiments, we report the detailed performance of all models alongside their standard deviations in Table~\ref{tab:main_results_std}. For all generative evaluations, the error bars (standard deviations) are calculated across $k$ independent sampling runs. Specifically, we compute the standard deviation over $k=16$ runs for AIME24, AIME25, and AMC23, and over $k=5$ runs for the remaining datasets.

\begin{table}[h]
    \centering
    \caption{\textbf{Detailed results with standard deviations.} This table corresponds to the main results presented, additionally reporting the standard deviations to demonstrate statistical significance. The standard deviations are calculated across independent sampling runs ($k=16$ for AIME24, AIME25, and AMC23, and $k=5$ for the others). The average summary columns are omitted here for brevity.}
    \label{tab:main_results_std}
    
    \resizebox{\textwidth}{!}{
        \begin{tabular}{l cccccc cc c}
            \toprule
            \multirow{2}{*}{\textbf{Model}} & \multicolumn{6}{c}{\textbf{In-domain Reasoning}} & \multicolumn{2}{c}{\textbf{OOD Reasoning}} & \textbf{General} \\
            \cmidrule(lr){2-7} \cmidrule(lr){8-9} \cmidrule(lr){10-10}
             & AIME24 & AIME25 & AMC23 & Math500 & Minerva & Olympia & GPQA-D & Connect4 & IFEval \\
            \midrule
            Qwen3-8b & 22.0$_{\pm 1.7}$ & 19.3$_{\pm 1.2}$ & 66.5$_{\pm 1.7}$ & 82.8$_{\pm 0.4}$ & 37.9$_{\pm 0.6}$ & 52.3$_{\pm 0.5}$ & 48.9$_{\pm 0.7}$ & 10.9$_{\pm 1.4}$ & 83.0$_{\pm 0.3}$ \\
            Qwen3-8b+SFT & 15.3$_{\pm 1.7}$ & 13.3$_{\pm 1.4}$ & 56.0$_{\pm 1.7}$ & 78.2$_{\pm 0.5}$ & 38.6$_{\pm 0.7}$ & 44.3$_{\pm 0.6}$ & 39.0$_{\pm 1.6}$ & 12.0$_{\pm 1.5}$ & 79.6$_{\pm 0.4}$ \\
            Qwen3-8b+RFT & 27.3$_{\pm 1.4}$ & 20.0$_{\pm 1.6}$ & 59.0$_{\pm 2.0}$ & 83.0$_{\pm 0.4}$ & 40.4$_{\pm 0.6}$ & 53.8$_{\pm 0.5}$ & 49.1$_{\pm 0.9}$ & 3.1$_{\pm 1.1}$ & 81.5$_{\pm 0.5}$ \\
            Qwen3-8b+\textbf{SFT+} & \textbf{29.3}$_{\pm 0.8}$ & 24.7$_{\pm 0.8}$ & 67.3$_{\pm 3.0}$ & 84.8$_{\pm 0.4}$ & \textbf{42.5}$_{\pm 0.5}$ & 54.5$_{\pm 0.5}$ & 47.3$_{\pm 1.2}$ & 14.1$_{\pm 1.3}$ & 80.0$_{\pm 0.6}$ \\
            Qwen3-8b\textbf{+\methodName} & 28.0$_{\pm 1.1}$ & \textbf{27.3}$_{\pm 1.1}$ & \textbf{71.5}$_{\pm 1.3}$ & \textbf{87.4}$_{\pm 0.5}$ & 39.7$_{\pm 0.5}$ & \textbf{58.5}$_{\pm 0.7}$ & 46.8$_{\pm 0.5}$ & \textbf{36.0}$_{\pm 1.8}$ & \textbf{84.8}$_{\pm 0.4}$ \\
            \bottomrule
        \end{tabular}
     }
\end{table}

\section{Experiment Setting}
\label{appd:exp_setting}

\textbf{Models and Datasets.}\quad We adopt Qwen3-8B and Llama3.1-8B-Instruct as the policy models for training. We intentionally choose instruction-tuned models over base models, since these already post-trained models are more susceptible to catastrophic forgetting \citep{lu2025onpolicydistillation}. If \methodName retains these models' prior knowledge during subsequent training, it validates the robustness of our approach. For Qwen3-8B, we enforce a non-thinking mode by explicitly inserting an empty thinking block (``\verb|<think>\n\n</think>\n\n|''). We leverage Gemini 2.5 Pro as the Oracle for rectification. 

Experiments are conducted using the English subset of DAPO-Math-17k \citep{yu2025dapo}. Leveraging our pipeline, we generate 4k contrastive pairs for Qwen3-8B. For Llama3.1-8B-Instruct model, we randomly sample 1.5k pairs, matching the number of correct responses the model could inherently generate, to ensure a fair comparison with Rejection sampling Fine-Tuning (RFT). All methods are compared using the same amount of training data.
\textbf{Evaluation.}\quad We evaluate performance across: \textit{in-domain reasoning}, including AIME24, AIME25, AMC23, Math500~\citep{hendrycks2measuring}, Minerva~\citep{lewkowycz2022solving}, and Olympia~\citep{he2024olympiadbench}; \textit{OOD reasoning}, including GPQA-Diamond (GPQA-D)~\citep{rein2024gpqa} and \texttt{Connect4}, and \textit{general instruction following} using IFEval~\citep{zhou2023instruction}. To strictly evaluate OOD reasoning without data contamination, we leverage GAMEBoT \citep{lin2025gamebot} to dynamically construct the \texttt{Connect4} dataset (see Appendix \ref{appd:connect4}). For reproducibility and fairness, we adopt the evaluation prompts from SoberEval \citep{hochlehnert2025soberreasoning}. We search for the optimal evaluation hyperparameters for the base models and apply them consistently across all compared methods.

\textbf{Hyperparameters.}\quad For data generation and evaluation, we employ a temperature of 0.7 and top-$p$ of 0.8, with a maximum token limit of 32,768. We train all models for 2 epochs with a context length of 8,192 and $\beta = 0.1$. To ensure a fair comparison, we systematically tune the hyperparameters for the baselines. Regarding learning rates, \methodName utilizes $1 \times 10^{-6}$ for Qwen3-8B and $2 \times 10^{-7}$ for Llama3.1-8B-Instruct. For SFT, RFT, and SFT+, we use learning rates of $5 \times 10^{-6}$ (Qwen3-8B) and $1 \times 10^{-6}$ (Llama3.1-8B-Instruct). The batch size is set to 32 for all the methods.

\textbf{RL Experiment Setting.}\quad For the RL experiment in \cref{sec:exper_rl}, we adopt GRPO~\citep{shao2024deepseekmath} implemented in \texttt{verl}~v0.6.0 with vLLM~v0.9.0 as the rollout backend. We use 4k English subset of DAPO-Math-17k for RL training. All runs are conducted on a single node with $8\times$ H800 (80GB) GPUs. For each prompt we draw a group of $n{=}8$ rollouts at temperature $1.0$. The training batch size is $256$ with a mini-batch size of $64$, and we train for $10$ epochs with a learning rate of $1\times10^{-6}$. For Qwen3-8B we keep the non-thinking mode by passing \texttt{enable\_thinking{=}False} to the chat template,
   consistent with the \methodName setup. We set the response length limit to 4096 to fit in memory.
\section{Connect4 Task}
\label{appd:connect4}
\textbf{Game Rules.}\quad Connect4 is a two-player game of perfect information played on a vertical grid of dimension $6 \times 7$. Players alternate turns dropping distinct pieces into one of the seven columns, where the piece occupies the lowest available row within that column due to simulated gravity. The objective is to form a contiguous line of four pieces either vertically, horizontally, or diagonally. It requires the LLM to \textit{parse the board state}, \textit{execute lookahead planning}, and \textit{identify winning topologies}. With a state-space complexity of approximately $4.5 \times 10^{12}$ legal positions, Connect4 offers a non-trivial, deterministic environment ideal for generating synthetic states to evaluate LLMs without data contamination.
\textbf{Dynamic Dataset Construction.}\quad To strictly evaluate OOD reasoning without the risk of data contamination in static benchmarks, we leverage the GAMEBoT framework \citep{lin2025gamebot}. While GAMEBoT supports full-game simulations, interactive evaluation is often inefficient and difficult to reproduce. Instead, we dynamically construct a static dataset of reasoning tasks rooted in \texttt{Connect4}. To ensure the model relies on symbolic reasoning, the board state is serialized into a text-based format (e.g., a list of coordinates).

The construction process involves:
\begin{itemize}

\item State Generation: We utilize two randomized agents to simulate gameplay, generating a diverse distribution of board states ranging from balanced positions to critical tactical scenarios.
\item Filtering and Balancing: We filter out duplicate states. Furthermore, to prevent LLMs from inflating scores by consistently predicting ``no winning moves'' (due to label imbalance), we downsample states with empty answers, maintaining them at a maximum ratio of 20\% of the dataset.
\item Dataset Compilation: For efficient evaluation, we construct a final dataset comprising 500 distinct instances.
\end{itemize}
\textbf{Evaluation Protocol.}\quad For each instance, LLMs are prompted to solve two problems in a query: 
\begin{itemize}
    \item \textit{Are there any potential winning moves to form 4-in-a-row for you? Output all winning moves.} 
    \item \textit{Are there any potential winning moves to form 4-in-a-row for your opponent? Output all winning moves.}
\end{itemize}
Ground truth is computed using a perfect solver within GAMEBoT. We parse the model's Chain-of-Thought (CoT) to extract the final answers and verify them against the game engine's ground truth.

\section{Additional Evaluation}
\label{appd:general_caps}
A central design goal of \methodName is to inject new knowledge without degrading the model's prior abilities. Beyond the benchmarks reported in the main text, we further evaluate on TruthfulQA and a 500-sample subset of MMLU-Pro (\cref{tab:general_caps}). The results suggest that on-policy rectification introduces useful knowledge without overwriting general competencies.

\begin{table}[ht]
\centering
\caption{General-capability evaluations on Qwen3-8B. \methodName preserves truthfulness and improves on MMLU-Pro.}\label{tab:general_caps}
\small
\begin{tabular}{lcc}
\toprule
Model & TruthfulQA & MMLU-Pro (500 subset) \\
\midrule
Qwen3-8B Base       & 68.39 & 51.2 \\
Qwen3-8B + \methodName  & 67.90 & 55.2 \\
\bottomrule
\end{tabular}
\end{table}

\section{Comparison with DPO+SFT}
\label{appd:dpo_sft}

We additionally compare \methodName against the DPO+SFT recipe~\citep{wang2024enhancing}, which combines preference optimization with supervised fine-tuning. Under the same setting on Qwen3-8B (\cref{tab:dpo_sft}), \methodName outperforms DPO+SFT on every benchmark.

\begin{table}[ht]
\centering
\caption{Comparison with DPO+SFT~\citep{wang2024enhancing} on Qwen3-8B. \methodName dominates across in-domain math, OOD reasoning, and instruction following.}\label{tab:dpo_sft}
\small
\begin{tabular}{lccccc}
\toprule
Model & AIME24 & AIME25 & AMC23 & Connect4 & IFEval \\
\midrule
Base                       & 22.0 & 19.3 & 66.5 & 10.9 & 83.0 \\
DPO+SFT~\citep{wang2024enhancing} & 24.6 & 18.9 & 66.5 & 21.9 & 81.5 \\
\methodName (Ours)         & \textbf{26.0} & \textbf{19.3} & \textbf{67.0} & \textbf{26.0} & \textbf{84.1} \\
\bottomrule
\end{tabular}
\end{table}

\section{Experiment on Qwen3-1.7B}
\label{appd:scale_general}

We additionally apply \methodName to Qwen3-1.7B (\cref{tab:qwen3_1p7b}). \methodName yields consistent gains across in-domain math and IFEval, indicating that the rectification-based recipe transfers to smaller capacities. \texttt{Connect4} is too difficult for both Base and \methodName at this scale, indicating that the model lacks the reasoning capacity for the task rather than reflecting a failure of the training recipe.

\begin{table}[ht]
\centering
\caption{\methodName on Qwen3-1.7B. The recipe transfers to smaller models.}\label{tab:qwen3_1p7b}
\small
\setlength{\tabcolsep}{4pt} 
\begin{tabular}{lccccc}
\toprule
Model & AIME24 & AIME25 & AMC23 & IFEval & Connect4 \\
\midrule
Base       & 12.67 & 9.33  & 41.00 & 67.64 & 0.00 \\
\methodName & 14.67 & 13.33 & 46.00 & 68.30 & 0.00 \\
\bottomrule
\end{tabular}
\end{table}




\clearpage

\end{document}